\documentclass[10pt,journal,compsoc]{IEEEtran}
\ifCLASSOPTIONcompsoc
  \usepackage[nocompress]{cite}
\else
  \usepackage{cite}
\fi
\ifCLASSINFOpdf
   \usepackage[pdftex]{graphicx}
   \graphicspath{{/}}
   \DeclareGraphicsExtensions{.pdf,.jpeg,.png}
\else
\fi
\usepackage{hyperref}
\usepackage{url}

\usepackage{color}

\begin{document}
%
\title{Learning and Tracking the 3D Body Shape of Freely Moving Infants from RGB-D sequences}
%
%
%
%

\author{Nikolas~Hesse,
        Sergi~Pujades,
        Michael~J.~Black,
        Michael~Arens,
        Ulrich~G.~Hofmann,
        and~A.~Sebastian~Schroeder
\IEEEcompsocitemizethanks{\IEEEcompsocthanksitem N. Hesse and M. Arens are with the Fraunhofer Institute of Optronics, System Technologies and Image Exploitation, Ettlingen, Germany.\protect\\
E-mail: nikolas.hesse@iosb.fraunhofer.de
\IEEEcompsocthanksitem S. Pujades and M. J. Black are with Max Planck Institute for Intelligent Systems, T\"ubingen, Germany.
\IEEEcompsocthanksitem U. G. Hofmann is with University Medical Center Freiburg, Faculty of Medicine, University of Freiburg, Germany.
\IEEEcompsocthanksitem A. S. Schroeder is with Ludwig Maximilian University, Hauner Children's Hospital, Munich, Germany.}
\thanks{Supplemental video available at \url{https://youtu.be/aahF1xGurmM}}
}

%
%

\newcommand{\smplb}{{SMPL$_\mathrm{B}$ }}
\newcommand{\nh}[1]{\textcolor{green}{\textbf{[NH: #1]}}}
\newcommand{\SP}[1]{\textcolor{blue}{\textbf{[SP: #1]}}}

\IEEEtitleabstractindextext{%
\begin{abstract}
Statistical models of the human body surface are generally learned from thousands of high-quality 3D scans in predefined poses to cover the wide variety of human body shapes and articulations.
Acquisition of such data requires expensive equipment, calibration procedures, and is limited to cooperative subjects who can understand and follow instructions, such as adults.
We present a method for learning a statistical 3D Skinned Multi-Infant Linear body model (SMIL) from incomplete, low-quality RGB-D sequences of freely moving infants.
Quantitative experiments show that SMIL faithfully represents the RGB-D data
and properly factorizes the shape and pose of the infants.
To demonstrate the applicability of SMIL, we fit the model to RGB-D sequences of freely moving infants
and show, with a case study, that our method captures enough motion detail for General Movements Assessment (GMA), a method used in clinical practice for early detection of neurodevelopmental disorders in infants.
SMIL provides a new tool for analyzing infant shape and movement and is a step towards an automated system for GMA. 
\end{abstract}

}

\maketitle

\IEEEdisplaynontitleabstractindextext

%
\IEEEpeerreviewmaketitle

\IEEEraisesectionheading{\section{Introduction}\label{sec:introduction}}

\IEEEPARstart{S}{tatistical} parametric models of the human body, such as SCAPE \cite{anguelov2005scape} or SMPL \cite{loper2015smpl}
describe the geometry of the body surface of an observed population in a low-dimensional space.
%
They are usually learned from dense high quality scans of the surface
of the human body.
Typically, subjects are instructed to stand in the same pose to
simplify the problem of modeling body shape.

Since the pioneering work of Blanz and Vetter
\cite{blanz1999morphable} on a morphable face model, parametric shape models have been evolved and have found a wide range of applications in computer vision and computer graphics.
For example, the low-dimensional representation of the human body
surface has played a  key role in enabling i) the precise capture of shape and pose of humans in motion from low quality RGB-D sequences \cite{bogo2015detailed};
ii) the temporal registration of highly dynamic motions of the human body surface \cite{bogo2017dynamic};
iii) the prediction of human shape and pose from single RGB images based on deep neural networks\cite{tan2017indirect,tung2017self,omran2018neural,pavlakos2018learning};
and iv) learning detailed avatars from monocular video \cite{alldieck2018detailed,alldieck2018video} .

Human movements contain key information allowing the infer of, for example, the performed task \cite{du2015hierarchical},
or internal properties of the observed subject \cite{lu2014human}.
In our work, we consider the application of {\it motion analysis}, i.e.~the acquisition and quantification of poses an observed subject strikes.
%
Human motion analysis is used in medicine for patient monitoring \cite{achilles2016patient}, quantifying therapy or disease progression~\cite{kontschieder2014quantifying}, or performance assessment \cite{clark2012validity}, e.g. by comparing the execution of a predefined movement with a reference motion \cite{coskun2018human}.
Most interestingly, it can be applied to the early detection of neurodevelopmental disorders like cerebral palsy (CP) in infants at a very early age.
The General Movements Assessment (GMA) approach enables trained experts to detect CP at an age of 2 to 4 months, based on assessing the movement quality of infants from video recordings \cite{prechtl1990qualitative}.
Infants with abnormal movement quality have very high risk of developing CP or 
minor neurological dysfunction~\cite{hadders2004general}.
While GMA is the most accurate clinical tool for early detection of
CP, it is dependent on trained experts and is consequently subject to
human perceptual variability. 
GMA experts require regular practice and re-calibration to assure
accurate ratings.
Automation of this analysis could reduce this variability and
dependence on human judgment.
To allow GMA automation,
a practical system must first demonstrate 
that it is capable of capturing the relevant information needed
for GMA.
Moreover, to allow its widespread use, 
the solution needs to be seamlessly integrated into the clinical routine.
Approaches aimed at GMA automation have relied on wearable sensors or vision-based systems for capturing infant motion.
For a review of existing methods, we refer the reader to \cite{marcroft2014movement} and \cite{hesse2018computer}.

\begin{figure}
\centering
\includegraphics[width=0.45\columnwidth]{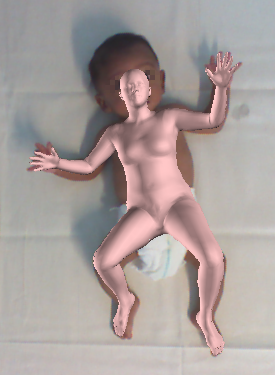}\llap{(a)}
\includegraphics[width=0.45\columnwidth]{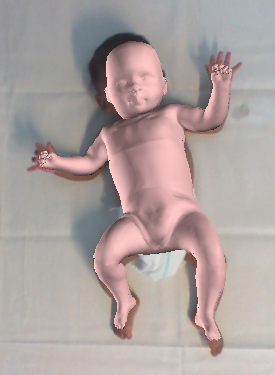}\llap{(b)}
\caption{(a) Simply scaling the SMPL adult body model and fitting it to an infant does not work as body proportions significantly differ. 
(b) The proposed SMIL model properly captures the infants' shape and pose.}
\label{fig:smpl_smil_fit}
\end{figure}

Inspired by previous work on capturing motion from RGB-D data using a body model \cite{bogo2015detailed},
we follow this direction in our work.
Two main problems arise on the way to capturing \textit{infant} motion using a body model. 
The first problem is that there is no infant body model. 
While parametric body models like SMPL \cite{loper2015smpl} cover a
wide variety of adult body shapes, the shape space does not
generalize to the new domain of infant bodies  (see Fig.~\ref{fig:smpl_smil_fit} a).
As the body part dimensions between infants and adults vary significantly,
the goal of this work is to learn an infant body model that faithfully captures the shape of infants (see Fig.~\ref{fig:smpl_smil_fit} b).

However, most statistical models are learned from high-quality scans, which are expensive, and demand cooperative subjects willing to follow instructions.
This is the second problem we face: 
there is no repository of high quality infant 3D body scans from which we could learn the statistics of infant body shape.
Acquiring infant shape data is not straightforward, as one needs to comply with strict ethics rules as well as a an adequate environment for the infants.
Therefore, we acquire sequences of moving infants in a children's hospital.
%
To record in a clinical environment, an acquisition system has to meet strict requirements.
We use one RGB-D sensor and a laptop as this provides a low-cost, easy-to-use alternative to bulky and expensive 3D scanners.
Our proposed system
produces minimal overhead to the standard examination protocol, and does not affect the behavior of the infants.
We use the captured RGB-D sequences to learn an infant body model.
%

Infant RGB-D data poses several challenges.
We have to deal with incomplete data, i.e.~partial views, where large parts of the body are occluded most of the time.
The data is of low quality and noisy, and captured subjects are not able to follow instructions and take predefined poses.
\textbf{Contributions.} We present the first work on 3D shape and 3D pose estimation of infants, 
as well as the first work on learning a statistical 3D body model from low-quality, incomplete RGB-D data of freely moving humans.
We contribute (i) a new statistical {\it Skinned Multi-Infant Linear} model (SMIL), learned from 37 RGB-D low-quality sequences of freely moving infants, and (ii) a method to register the SMIL model to the RGB-D sequences, capable of handling severe occlusions and fast movements. 
Quantitative experiments show how SMIL properly factorizes the pose and the shape of the infants,
and allows the captured data to be accurately represented in a low-dimensional space.
With a case-study involving a high-risk former preterm study population,
we demonstrate that the amount of motion detail captured by SMIL is
sufficient to enable accurate GMA ratings by humans.
Thus, SMIL provides a fundamental tool that can form a component in an automated system
for the assessment of GMs.
We make SMIL available to the community for research purposes at \url{http://s.fhg.de/smil}.
This article is an extended version of \cite{hesse2018learning}.

\section{Related Work}
We review two main areas: 
the creation of statistical models of the human body surface and 
the estimation of shape and pose of humans in movement from RGB-D sequences.

\subsection{Human Surface Models}

Statistical parametric models of the human body surface are usually based on the {\it Morphable Model} idea \cite{blanz1999morphable}, stating that a single surface representation can morph and explain the different samples in a population.
These models can be intuitively viewed as a mathematical function taking shape and pose parameters as input and returning a surface mesh as output.
The {\it shape space} and the {\it pose prior}, i.e. the statistics of most plausible poses,
are learned by registering the single surface representation, i.e. a template surface, to real-world data.
The shape space and the pose prior allow a compact representation of the human body surface describing the geometry of the human body surface of an observed population in a low-dimensional space.
Existing models have been learned from different real-world data.
For example, to model the variation of faces De-Carlo et al.~\cite{decarlo1998anthropometric}  learn a model from a cohort of anthropometric measurements.
Blanz and Vetter \cite{blanz1999morphable}, use dense geometry and color data to learn their face model.
Allen et al.~\cite{allen2003space} use the CAESAR dataset to create {\it the space of human body shapes} by using the geometry information as well as sparse landmarks identified on the bodies.
%
Similarly, Seo et Magnenat-Thalmann \cite{seo2003automatic} 
learn a shape space from high quality range data. 
Angelov et al. propose SCAPE~\cite{anguelov2005scape}, a statistical body model learned from high quality scan data,
which does not only contain the shape space, but also accounts for the pose dependent deformations, i.e. the surface deformations that a body undergoes when different poses are taken.
SCAPE and all successive statistical models of the human body surface \cite{hasler2009statistical, hirshberg2012coregistration, freifeld2012lie, chen2013tensor, zuffi2015stitched, loper2015smpl, pishchulin2017building}
or their soft tissue dynamics \cite{pons2015dyna, loper2015smpl, Meekyoung:siggraph}
have been learned from a relatively large number of high quality range scans of adult subjects.
Adults, in contrast to infants, are typically cooperative and can be instructed to strike specific poses during scanning.

Animal shape modeling methods face a similar difficulty as ours: live animals are generally difficult to instruct and their motions make them difficult to scan.
Thus, they provide a source of inspiration to create models without a large cohort of high quality scans.
Cashman and Fitzgibbon ~\cite{cashman2013shape}  learn a deformable model of dolphins by using manually annotated 2D images.
Kanazawa et al.~\cite{kanazawa2016learning} learn the deformations and the stiffness of the parts of a 3D mesh template of an animal from manually annotated 2D images.
To create the animal model SMAL~\cite{zuffi20173d}, Zuffi et al. circumvent the difficulty to instruct and scan real animals by using a small dataset of high quality scans of toy figurines. 
The SMAL model can be fit to new animals using a set of multi-view images with landmarks and silhouette annotations.

In this work, we learn our statistical model of the shape of infants from low-quality RGB-D sequences.
While there are several methods that fit body shape to RGB-D data, we do not know of any that estimates a statistical model from such input.
Our method does not rely on manually annotated landmarks and leverages the ones that are automatically extracted from RGB images \cite{wei2016cpm, cao2017realtime, simon2017hand}.

\subsection{Capturing motion from RGB-D using a body model}
The existing body models have proven to be successful in capturing the pose and shape of a subject from RGB-D sequences.
The model is parametrized with the shape, giving information about the joint locations, and the pose, defining the angles between the limbs. Once a model is registered to the input data, one can obtain the desired motion information.
As the joint locations depend on the shape, the closer the estimated body shape is to the actual subject's shape, the better, i.e. more accurate, the tracking of motions will be.

Parametric body models generally model the space of human bodies without clothing or hair.
Some approaches propose to overcome the discrepancy between such a shape space and the real world by creating personalized avatars from the input data and then registering them to the dynamic sequences by keeping the personalized shape fixed.
This creates an additional step and usually requires cooperative subjects to take predefined poses.
%
%
For example, personalized avatars are created from multiple Kinect scans \cite{weiss2011home,helten2013personalization,zheng2014scape}, Kinect fusion scans \cite{zhang2014quality,chen2016realtime}
or laser scanners \cite{ye2012performance}
and the obtained avatars are then registered to different RGB-D scans of the same person in different poses.
%






Other methods use a parametric body model to capture the pose without the preliminary step of the personalized shape creation.
Ganapathi et al.~\cite{ganapathi2012real} use a simplistic body model for real-time pose tracking from range data.
Ye and Yang \cite{ye2014real} introduce a method for real-time shape and pose tracking from depth.
They register an articulated deformation model to point clouds within a probabilistic framework.
Chen et al.~\cite{chen2013tensor} capture shape and pose from Kinect data using a tensor-based body model.
Yu et al.~\cite{yu2017bodyfusion} introduce an approach for real-time reconstruction of non-rigid surface motion from RGB-D data using skeleton information to regularize the  shape deformations.
They extend this approach by combining a parametric body model to represent the inner body shape with a freely deformable outer surface layer capturing surface details \cite{yu2018doublefusion}.

Bogo et al.~\cite{bogo2015detailed} fit a multi-resolution body model to Kinect sequences.
This work is the closest to ours, which is why we give a brief summary and identify similarities and differences.
Bogo et al.~aim at creating highly realistic textured avatars from RGB-D sequences.
They capture shape, pose and appearance of freely moving humans based on a parametric body model, which is learned from 1800 high-quality 3D scans of 60 adults.
They create a personalized shape for each sequence, by accumulating shape information over the sequence in a ``fusion cloud''.
The captured subjects wear tight clothing and take a predefined pose at the beginning of each sequence, as is common in scanning scenarios.
They use their body model at different resolutions in a coarse-to-fine manner to increasingly capture more details.
They use a displacement map to represent fine details that lie beyond the resolution of the model mesh.
The model contains a head-specific shape space for retrieving high-resolution face appearance.
%

In our work, we also capture shape and pose of sequences containing unconstrained movements.
To create a personalized shape, we also merge all temporal information into fusion clouds.
In the gradient-based optimization, some of our energy terms are similar to the ones from \cite{bogo2015detailed}.
In contrast to their work, an initial infant body model is not available and we must create it. 
We adapt an existing adult body model to a different domain, namely infants, to use as an initial model for registering our infant sequences.
The fact that infants lie in supine position in our scenario presents two different constraints.
First, it means that very few backs are visible and we have to deal with large areas of missing data in our fusion clouds.
Second, as infants are in contact with the background, i.e. the examination table,
we can not rely on a background shot to segment the relevant pointcloud.
When the infants move they wrinkle the towel they are lying on with their hands and feet.
However, we can take advantage of the planar geometry to fit a plane to the table data to segment it.
Moreover,  we can (and do) use the fitted plane as a geometric constraint, 
as we know the back of the infants can not be inside the examination table.
Also, in contrast to Bogo et al.~\cite{bogo2015detailed}, we can not rely on predefined poses for initialization since the infants are too young to strike poses on demand. We contribute a new automatic method for choosing the best poses for initialization.
Moreover, the clothing in our setting is not constrained: we have to deal with diapers, onesies and tights. 
In particular, diapers pose a challenge since their shape largely deviates from the human body. We handle the unconstrained cloth condition by segmenting the points corresponding to clothes, and by introducing different energy terms for clothing and skin parts.
Finally, in our work we do not use the appearance of the surface, but rather use 
the RGB information to extract 2D landmark estimates to have individual constraints on the face and hand rotations.

\section{Learning the Skinned Multi-Infant Linear model from low quality RGB-D data}
\begin{figure*}[t]
\includegraphics[width=\textwidth]{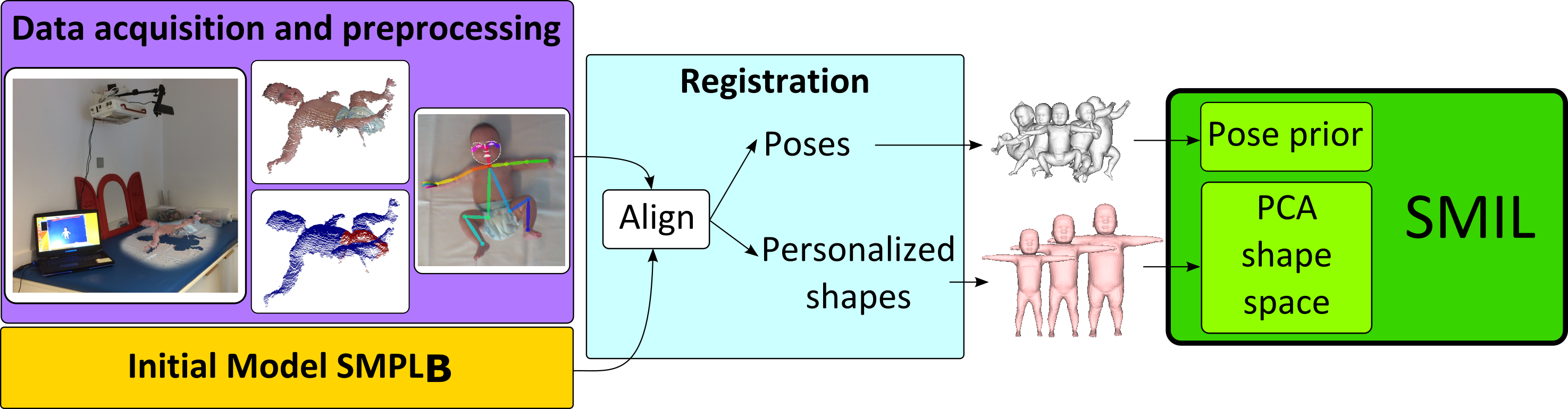}
\caption{Skinned Multi-Infant Linear (SMIL) model creation pipeline. We
  create an initial infant model based on SMPL. We perform background
  and clothing segmentation of the recorded sequences in a
  preprocessing step, and estimate body, face, and hand landmarks in
  RGB images. We register the initial model, SMPL$_\mathrm{B}$, to the
  RGB-D data, and create one personalized shape for each sequence,
  capturing infant shape details outside the SMPL shape space. We
  learn a new infant specific shape space by performing PCA on all
  personalized shapes, with the mean shape forming our base template.
We also learn a prior over plausible poses from a sampled subset of
all poses. }
\label{fig:pipe}
\end{figure*}
Learning a body model from data is a chicken-and-egg problem.
We need a model to register the data to a common topology, and we need
registered, or aligned, meshes to learn a model.
Since no infant body model is available, we first create an initial
infant model by adapting the adult SMPL model \cite{loper2015smpl} (see Sec. \ref{sec:inmo}).
We then register this initial model to RGB-D sequences of moving infants (Sec. \ref{sec:reg}).
To mitigate the incompleteness of data due to the monocular setup,
we accumulate shape information from each sequence into one ``personalized shape'' (Sec. \ref{sec:pers}).
Finally, we learn a new infant shape space from all personalized
shapes, as well as a new prior over plausible infant poses from our registrations (Sec. \ref{sec:learn}).
An overview of the complete learning pipeline is given in Fig.~\ref{fig:pipe}.

\subsection{Data}
\label{sec:data}
There are multiple reasons why no public repository of infant 3D scans exists.
Protection of privacy of infants is more strict as compared to adults.
The high cost of 3D scanners prevents them from being widespread.
Creating a scanning environment that takes into consideration the
special care required by infants, like warmth and hygiene, requires
additional effort.
Finally, infants can not be instructed to strike poses on demand, which is usually required in standard body scanning protocols.
RGB-D sensors offer a cheap and lightweight solution for scanning infants, only requiring the sensor and a connected laptop.
The data used to learn SMIL was obtained by setting up recording stations at a children's hospital where infants and parents regularly visit for examinations.
The acquisition protocol was integrated in the doctor's medical
routine in order to minimize overhead.

\textbf{Preprocessing.}
We transform depth images to 3D point clouds using the camera calibration.
To segment the infant from the scene, we fit a plane to the background table of the 3D point cloud using RANSAC \cite{fischler1981random} and remove all points close to or below the table plane and apply a simple cluster-based filtering.
Further processing steps operate on this segmented cloud, in which only points belonging to the infant remain.
Plane-based segmentation is not always perfect, e.g. in case of a wrinkled towel very close to the infant body, some noise may remain.
However, the registration methods have proven to be robust to outliers
of this kind.
The estimated table plane will be reused for constraining the infants' backs in the registration stage (Sec.~\ref{sec:reg}).


In order to avoid modeling diapers and clothing wrinkles in the infant shape space, we
segment the input point clouds into {\it clothing} and {\it skin} using the color information 
by adapting the method from Pons-Moll et al.~\cite{pons2017clothcap}.
We start by registering the initial model to one scan and perform an unsupervised k-means clustering to obtain 
the dominant modes in RGB. We manually define the clothing type  to be: naked, diaper, onesie long, onesie short or tights.
This determines the number of modes and the cloth prior.
The dominant modes are used to define probabilities for each 3D point being labeled as cloth or skin.
We transfer the points' probabilities to the model vertices, and solve a minimization problem on a Markov random field
defined by the model topology.
We transfer the result of the model vertices to the original point cloud, and we obtain a clean segmentation 
of the points belonging to clothing (or diaper) and the ones belonging to the skin.
An example of the segmentation result can be seen in the {\it Data
  acquisition and preprocessing} box of Fig.~\ref{fig:pipe}; blue is
skin and red is diaper.
To avoid registering all scans twice, i.e. a first rough registration to segment the sequence and a second to obtain the final registration, we transfer the clothing labels from the registration at frame $t-1$ to the point could at frame $t$.
In practice this works well, since body changes in consecutive frames are relatively small.

Scanning of adults typically relies on them striking a simple pose to
facilitate model fitting and registrations.
The scanned infants can not take a predefined pose to facilitate an initial estimate of model parameters.
However, existing approaches on 2D pose estimation from RGB images (for adults) have achieved impressive results.
Most interestingly, experiments show that applying these methods to
images of infants produces accurate estimates of 2D pose~\cite{hesse2018computer}.
In order to choose a ``good'' candidate frame to initialize the model parameters (see Sec.~\ref{sec:init}), we leverage the 2D body landmarks together with their confidence values. 
From the RGB images we extract body pose \cite{cao2017realtime} as well as face \cite{wei2016cpm} and hand \cite{simon2017hand} landmarks. We experimentally verify that they provide key information on the head and hand rotations to the registration process, which is complementary to the noisy point clouds.

\subsection{Initial model}
\label{sec:inmo}
We manually create an initial model by adapting the {\it Skinned Multi-Person Linear model} (SMPL) \cite{loper2015smpl}, which we briefly recap.
SMPL is a linear model of shape and pose.
It represents shape deformations as a combination of identity specific shape (shape blend shapes) and pose-dependent shape (pose blend shapes).
The pose blend shapes are learned from 1786 registered scans of adults in predefined poses, while the 
shape blend shapes are learned from registered scans of 1700 males and 2100 females from the CAESAR data set \cite{robinette2002civilian}.
The SMPL shape space is represented by a mean template shape and
principal shape directions that are created by performing principal
component analysis (PCA) on pose-normalized registered meshes.
The shape of a body is describe by a vector of linear coefficients,
$\beta$, that multiply the principal component displacements.
The model can be viewed as a mapping from shape and pose parameters to a shaped and posed mesh.
Shape and pose blend shapes are modeled as vertex offsets, which are added to the mean template, and the result is then transformed by a standard blend skinning function to form the output mesh.
The SMPL template consists of 6890 vertices and 23 body joints. Each
body joint has 3 degrees of freedom (DoF) resulting, with 3 DoF for
global rotation, in 72 pose parameters, $\theta$.
SMPL contains a learned joint regressor for computing joint locations from surface vertices.

\textbf{Adaptation to infants.}
We manually create an initial infant mesh using makeHuman, an open
source software for creating 3D characters.
We wish to register SMPL to this mesh to use this base shape in the
SMPL model.
Directly registering SMPL to the infant mesh fails due to differences in size and proportions.
We make use of the fact that meshes exported from makeHuman share the same topology, independent of shape parameters.
We register SMPL to an adult makeHuman mesh, and
describe makeHuman vertices as linear combinations of SMPL vertices.
This allows us to apply this mapping to the infant mesh and transfer it to the SMPL topology.
We then replace the SMPL base adult-shape template with the registered infant mesh.

We further scale the SMPL pose blend shapes, which correct skinning
artifacts and pose-dependent shape deformations, to infant size.
Specifically, we divided infant height by average adult height and
multiply the blend shapes by this factor.
We keep the SMPL joint regressor untouched, since we found that it worked well for infants in our experiments.
As SMPL pose priors, i.e. prior probabilities of plausible poses, are learned from data of adults in upright positions,
these can not be directly transferred to lying infants.
We manually adjust them experimentally.
Specifically we penalize bending of the spine since the infants are
lying on their backs. 
Without this penalty, the model tries to explain shape deformations with pose parameters.

\subsection{Registration}
\label{sec:reg}
We register the initial model to the segmented point cloud using gradient-based optimization.
%
The main energy being optimized w.r.t.\ shape $\beta$ and pose $\theta$ parameters is
\begin{equation}
\label{eq:error_shape_pose}
\mathrm{E(\beta, \theta)} =\mathrm{E_{data}} + \mathrm{E_{lm}} + \mathrm{E_{table}} + \mathrm{E_{sm}} + \mathrm{E_{sc}} + \mathrm{E_{\beta}} + \mathrm{E_{\theta}},
\end{equation}
where the weight factors $\lambda_x$ associated with term $E_x$ are omitted for compactness.
In the following, we explain each term of the energy in detail. 

\textbf{Data term.}
The data term $\mathrm{E_{data}}$ consists of two different terms:
\begin{equation}
\label{eq:data}
\mathrm{E_{data}}=  \mathrm{E_{s2m}} + \lambda_{\mathrm{m2s}}\mathrm{E_{m2s}}.
\end{equation}

$\mathrm{E_{s2m}}$ accounts for the distance of the scan points to the model surface and
$\mathrm{E_{m2s}}$ accounts for the distance of the visible model vertices to the scan points.

$\mathrm{E_{m2s}}$ can be written as
\begin{equation}
\mathrm{E_{m2s}}(M, P)  = \sum_{m_i\in \mathrm{vis(M)}}{\rho(\min_{v\in P}||(m_i, v))||)},
\end{equation}
where $M$ denotes the model surface and 
$\rho$ is the robust Geman-McClure function \cite{GemanMcClure1987}.
We denote the scan points as $P$. In the preprocessing stage, $P$ is segmented into
the scan points belonging to the skin ($P_{\mathrm{skin}}$) and the ones belonging to clothing ($P_{\mathrm{cloth}}$).
The function $\mathrm{vis(M)}$ selects the 
visible model vertices. The visibility is computed using the Kinect V1 camera calibration and the OpenDR renderer \cite{loper2014opendr}.

$\mathrm{E_{s2m}}$ consists of two terms,
\begin{equation}
\label{eq:s2m}
\mathrm{E_{s2m}}=  \lambda_{\mathrm{skin}}\mathrm{E_{skin}} + \lambda_{\mathrm{cloth}}\mathrm{E_{cloth}}.
\end{equation}
$\mathrm{E_{skin}}$ enforces the skin points to be close to the model mesh and
$\mathrm{E_{cloth}}$ enforces the cloth points to be outside the model mesh.

%
The skin term can be written as 
\begin{equation}
\mathrm{E_{skin}}(M, P_{\mathrm{skin}}, W)  = \sum_{v_i\in P_{\mathrm{skin}}}{W_i\rho(\mathrm{dist}(v_i, M))},
\label{eq:skin}
\end{equation}
where $W$ are the skin weights.
The cloth term is divided into two more terms, depending on cloth points lying inside or outside the model mesh:
\begin{equation}
E_{cloth} = E_{outside} + E_{inside},
\end{equation}
with
\begin{equation}
\mathrm{E_{outside}}(M, P_{\mathrm{cloth}}, W)  = \sum_{v_i\in P_{\mathrm{skin}}}{\delta_i^{\mathrm{out}}W_i\mathrm{dist}(v_i, M)^2},
\end{equation}
where $\delta_i^{\mathrm{out}}$ is an indicator function, returning 1 if $v_i$ lies outside the model mesh, else 0 (Eq. 3 from \cite{zhang2017detailed}),
and
\begin{equation}
\mathrm{E_{inside}}(M, P_{\mathrm{cloth}}, W)  = \sum_{v_i\in P_{\mathrm{skin}}}{\delta_i^{\mathrm{in}}W_i\rho(\mathrm{dist}(v_i, M))},
\end{equation}
with $\delta_i^{\mathrm{in}}$ an indicator function, returning 1 if $v_i$ lies inside the model mesh, else 0.

\textbf{Landmark term.}
Due to the low-quality of depth data, depth-only methods can not reliably capture details like head or hand rotations.
However, we can estimate 2D landmark positions from the RGB images and use them as additional constraints in the optimization energy.
Body landmarks \cite{cao2017realtime} are used for initialization (Sec.~\ref{sec:init}), whereas face \cite{wei2016cpm} and hand \cite{simon2017hand} landmarks are used in the registration energy of every frame.
In the cases where the face detection \cite{wei2016cpm} fails, mostly profile faces, 
we use the ears and eyes information from the body pose estimation method \cite{cao2017realtime}.
These help to guide the head rotation in these extreme cases.

The landmark term $\mathrm{E_{lm}}$ is similar to Eq. 2 from \cite{bogo2016keep}, where the distances between the 2D landmarks estimated  from RGB and the corresponding projections of the 3D model landmarks are measured. 
Instead of using the body joints, we only use the estimated 2D face landmarks (nose, eyes outlines, mouth outline and ears) as well as the hand landmarks (knuckles). 
We note the set of all markers as $\mathrm{L}$.
%
%
The 3D model points corresponding to the above landmarks were manually selected through visual inspection.
They are projected into the image domain using the camera calibration matrix
in order to compute the final 2D distances to the estimated landmarks.

The landmark term is then
\begin{equation}
\mathrm{E_{lm}} = \lambda_{\mathrm{lm}}\sum_{l \in \mathrm{L}}{c_l \rho(|| l_{M} - l_{{\mathrm{est}}} ||)},
\end{equation}
where $c_l$ denotes the confidence of an estimated landmark 2D location $l_{{\mathrm{est}}}$, and $l_{M}$ is the projected model landmark location.
All confidences from the different methods are in the interval $[0, 1]$, making them comparable in terms of magnitudes.

\textbf{Table term.}
The recorded infants are too young to roll over, which is why the back is rarely seen by the camera.
However, the table on which the infants lie, lets us infer shape information of the back.
We assume that the body can not be inside the table, and that a large part of the back will be in contact with it.
We note the table plane as $\Pi$.
The table energy has two terms: $\mathrm{E_{in}}$ prevents the model vertices $M$ from lying inside the table (i.e. behind the estimated table plane),
by applying a quadratic error term on points lying inside the table.
$\mathrm{E_{close}}$ acts as a gravity term, by pulling the model vertices $M$ which are close to the table towards the table,
by applying a robust Geman-McClure penalty function to the model points which are close to the table.

We write the table energy term as 
\begin{equation}
\mathrm{E_{table}} =  \lambda_\mathrm{{in}}\mathrm{E_{in}} + \lambda_{\mathrm{close}}\mathrm{E_{close}},
\end{equation}
with
\begin{equation}
\mathrm{E_{in}(M)} =\sum_{x_i \in M}\delta_{i}^{\mathrm{in}}(x_i){\mathrm{dist}(x_i,\Pi)^2},
\end{equation}
and
\begin{equation}
\mathrm{E_{close}(M)} = \sum_{x_i \in M}\delta_{i}^{\mathrm{close}}(x_i)\rho(\mathrm{dist}(x_i,\Pi)),
\end{equation}
where $\delta_{i}^{\mathrm{in}}$ is an indicator function, returning 1 if $x_i$ lies inside the table (behind the estimated table plane), or 0 otherwise
and $\delta_{i}^{\mathrm{close}}$ is an indicator function, returning 1 if $x_i$ is close to the table (distance less than 3 cm) and faces away from the camera, or 0 otherwise.

To account for soft tissue deformations of the back, which are not modeled, we allow the model to virtually penetrate the table.
We effectively enforce this by translating the table plane by 0.5 cm,
i.e.~pushing the virtual table back.

The weight of the table term needs to be balanced with the data term to avoid a domination of the gravity term, 
keeping the body in contact with the table while the data term suggests otherwise.

\textbf{Other terms.}
Depth data contains noise, especially around the borders. To avoid jitter in the model caused by that noise, we add a temporal pose smoothness term.
It avoids important changes in pose unless one of the other terms has strong evidence.
The temporal pose smoothness term $\mathrm{E_{sm}}$ is the same as in
Eq. 21 in \cite{romero2017embodied} 
and penalizes large differences between the current pose $\theta$ and the pose from the last processed frame $\theta'$.
The penalty for model self intersections $\mathrm{E_{sc}}$
and the shape prior term $\mathrm{E}_\beta$ are the same as in Eq. 6 and Eq. 7 in \cite{bogo2016keep} respectively.
Bending the model in unnatural ways might decrease the data term error, which is why the pose prior term keeps the pose parameters in a realistic range.
The SMIL pose prior consists of a mean and covariance matrix that were
learned from 37,000 sample training poses; these are not used during testing.
$\mathrm{E}_\theta$ penalizes the squared Mahalanobis distance between $\theta$ and the pose prior, as described in \cite{bogo2015detailed}.

\subsection{Registration Optimization}
To compute the registrations of a sequence
we start by computing an initial shape using 5 frames.
In this first step, we optimize for the shape and pose parameters, $\beta$ and $\theta$, as well as the global translation $t$.
The average shape parameters from these 5 frames will be kept fixed and used later on as a shape regularizer.
Experiments showed that otherwise the shape excessively deforms in order to explain occlusions in the optimization process.

With the initial shape fixed, we compute the poses for all frames in the sequence,
i.e. we optimize the following energy w.r.t. pose parameters $\theta$ and the global translation $t$:
\begin{equation}
\mathrm{E(\theta, t)} =\mathrm{E_{data}} + \mathrm{E_{lm}} + \mathrm{E_{table}} + \mathrm{E_{sm}} + \mathrm{E_{sc}} + \mathrm{E_{\theta}}.
\label{eq:error_poseonly}
\end{equation}
Notice that this energy is equal to Eq. \ref{eq:error_shape_pose} without the shape term $\mathrm{E_{beta}}$, as shape is kept fixed.
We denote $S_f$ the computed posed shape at frame $f$.

In the last step, we compute the registration meshes $R_f$
and allow the model vertices $v \in R_f$ to freely deform to best explain the input data.
We optimize w.r.t. $v$ the energy
\begin{equation}
\mathrm{E}(v) = \mathrm{E_{data}} + \mathrm{E_{lm}} + \mathrm{E_{table}} + \mathrm{E_{cpl}},
\label{eq:error_cpl}
\end{equation}
where
$\mathrm{E_{cpl}}$ is a ``coupling'' term, used to keep the registration edges close to the edges of the initial shape. 
We use the same energy term as Eq. 8 from \cite{bogo2015detailed}
\begin{equation}
\mathrm{E_{cpl}}(R_f, S_f) = \lambda_{cpl} \sum_{e \in V'}{||(AR)_e - (AS)_e||_F^2},
\end{equation}
where $V'$ denotes the edges of the model mesh.
$AR$ and $AS$ are edge vectors of the triangles of $R_f$ and $S_f$,
and $e$ indexes the edges.
The results of these optimizations are the final registrations.

\begin{figure}[t]
	\centering
        \includegraphics[width=0.45\textwidth]{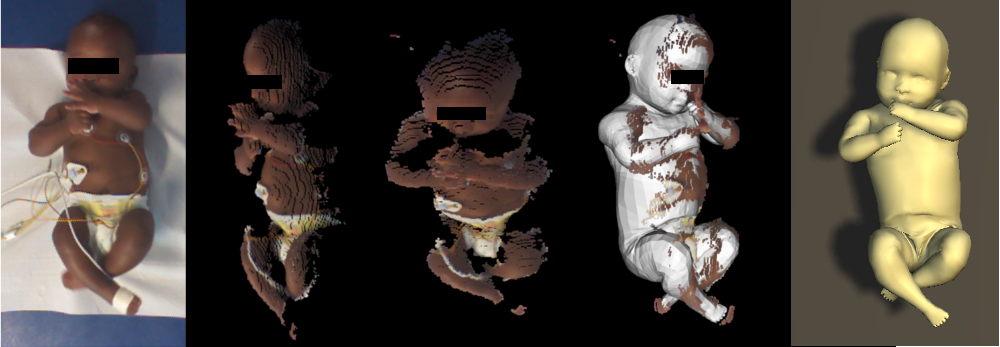}
	\includegraphics[width=0.45\textwidth]{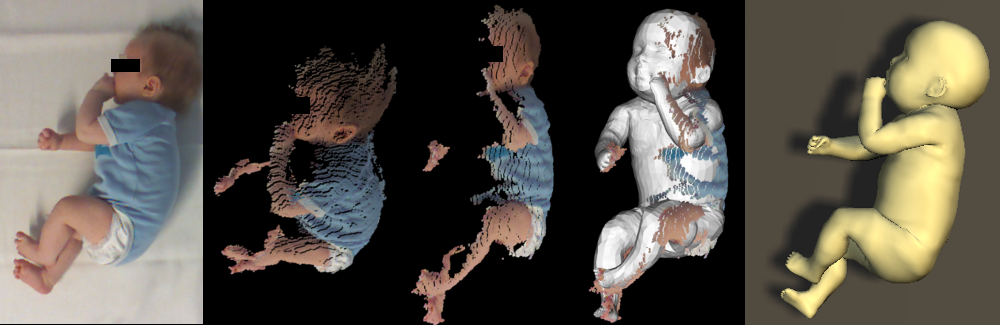}
	\includegraphics[width=0.45\textwidth]{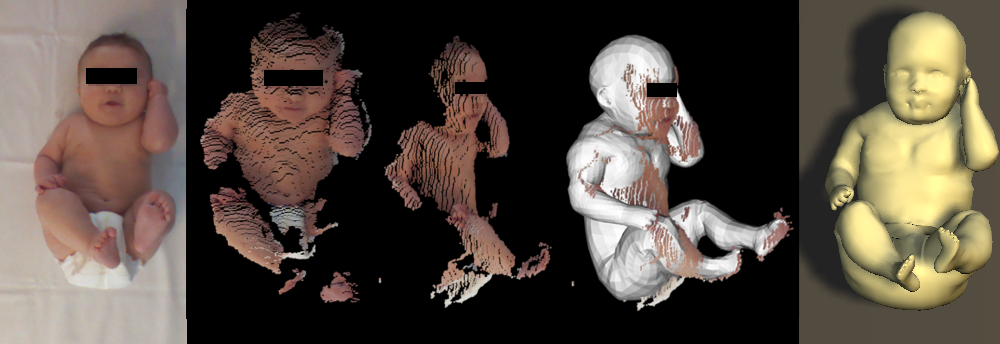}
    \caption{Registrations. From left to right: RGB, point cloud, point cloud (other view), point cloud with registered SMIL, rendered registration.}
\label{fig:alignsample1}
\end{figure}

All energies are minimized using a gradient-based dogleg minimization method~\cite{nocedal2006numerical} with 
OpenDR \cite{loper2014opendr} and Chumpy \cite{loperchumpy}.
We display registration samples in Fig.~\ref{fig:alignsample1}.

\subsection{Initialization}
\label{sec:init}
In order to find the global minimum, the optimization needs a good initial estimate.
In adult settings, subjects are usually asked to take an easy pose, e.g. T-pose (extended arms and legs), at the start of the recording.
Infants are not able to strike poses on demand, which is why we can not rely on a predefined pose.

We automatically find an initialization frame containing an ``easy'' pose by relying on 2D landmark estimates acquired in the preprocessing stage.
We make the assumption that a body segment is most visible if it has maximum 2D length over the complete sequence.
Perspective projection would decrease 2D body segment length and therefore visibility.
The initialization frame is chosen as 
\begin{equation}
f_{\mathrm{init}} = \mathrm{argmax}_f \sum_{s \in S} \mathrm{len}(s, f) * c(s, f),
\label{eq:init}
\end{equation}
where $S$ is the set of body segments, $\mathrm{len}(s, f)$ is the 2D length of the segment $s$ in frame $f$, and $c(s, f)$ is the estimated confidence of joints belonging to $s$ in frame $f$.

For $f_{\mathrm{init}}$, we optimize a simplified version of Eq.~\ref{eq:error_shape_pose}, i.e. the initialization energy

\begin{equation}
\label{eq:init}
\mathrm{E_{init}} = \lambda_{\mathrm{j2d}}\mathrm{E_{j2d}} + \lambda_{\theta}\mathrm{E_{\theta}} + \lambda_{\mathrm{a}}\mathrm{E_a} + \lambda_{\beta}\mathrm{E_{\beta}} + \lambda_{\mathrm{s2m}}\mathrm{E_{s2m}}
\end{equation}
where $\mathrm{E_{j2d}}$ is similar to $\mathrm{E_{lm}}$ with landmarks being 2D body joint positions. $\mathrm{E_\theta}$ is a strong pose prior, $\mathrm{E_a(\theta)} = \sum_i{\exp(\theta_i)}$ is an angle limit term for knees and elbows and $\mathrm{E_\beta}$ a shape prior.
Its minimum provides a coarse estimation of shape and pose, which is refined afterwards.
In contrast to \cite{bogo2016keep}, we omit the self intersection term, and add a scan-to-mesh distance term $\mathrm{E_{s2m}}$, containing 3D information, while \cite{bogo2016keep} solely relies on 2D information.

\subsection{Personalized Shape}
\label{sec:pers}
To capture the subject specific shape details, we create one personalized shape from each sequence,
which we do not restrict to the shape space of the model.
We \textit{unpose} a randomly selected subset of 1000 frames per sequence. The process of unposing changes the model pose to a normalized pose (T-pose) in order to remove variance related to body articulation.
For each scan point, we calculate the offset normal to the closest model point. After unposing the model, we add these offsets to create the unposed point cloud for each of the 1000 frames.
Since the recorded infants lie on their backs most of the time, the unposed clouds have missing areas on the back side.
To take advantage of the table constraint in each frame and sparsly fill the missing areas, we add  {\it virtual points}, i.e.~ points from model vertices that belong to faces oriented away from the camera, to the unposed cloud.
We retain the clothing segmentation labels for all unposed scan points.
We call the union of all unposed point clouds including virtual points
the {\it fusion cloud}; cf.~\cite{bogo2015detailed}.

To compute the personalized shape, we uniformly sample 1 million
points at random from the fusion cloud and proceed in two stages.
First, we optimize 
$\mathrm{E} = \mathrm{E_{data}} + \mathrm{E_{\beta}}$
w.r.t.~the shape parameters $\beta$, and keep the pose $\theta$ fixed in the zero pose of the model (T-pose with legs and arms extended).
We obtain an initial shape estimate that lies in the shape space of the initial model SMPL$_\mathrm{B}$.
Second, we allow the model vertices to deviate from the shape space, 
but tie them to the shape from the first stage with a coupling term.
We optimize $\mathrm{E} = \mathrm{E_{data}} + \mathrm{E_{cpl}}$ w.r.t.~the vertices.

The clothing segmentation is also transformed to the unposed cloud and
therefore, the fusion cloud is labeled into clothing and skin
parts. These are used in the data term to enforce that the clothing points to lie outside the model surface and to avoid learning clothing artifacts in the shape space.

\subsection{Learning SMIL shape space and pose prior}
\label{sec:learn}
\begin{figure}[t]
	\centering
        \includegraphics[width=0.15\textwidth]{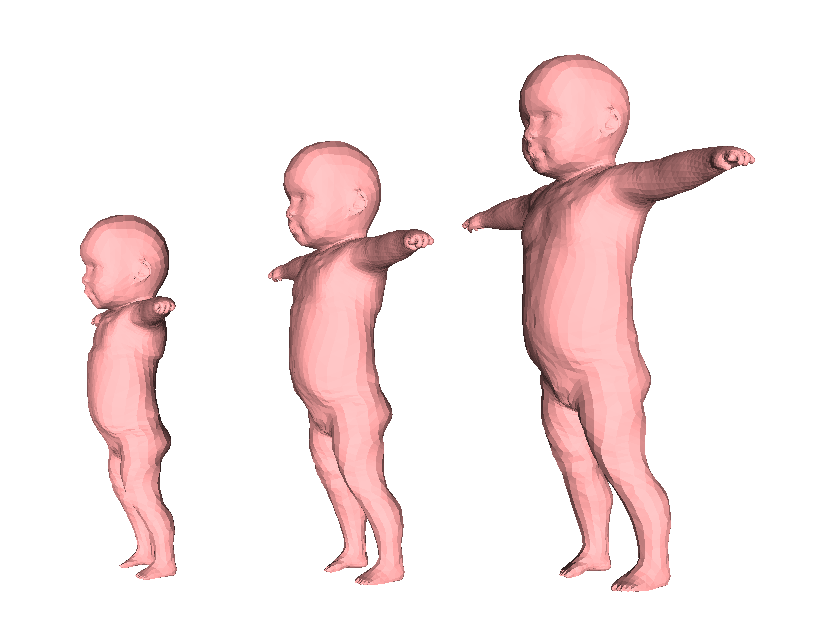}\llap{sc 1}
	\includegraphics[width=0.15\textwidth]{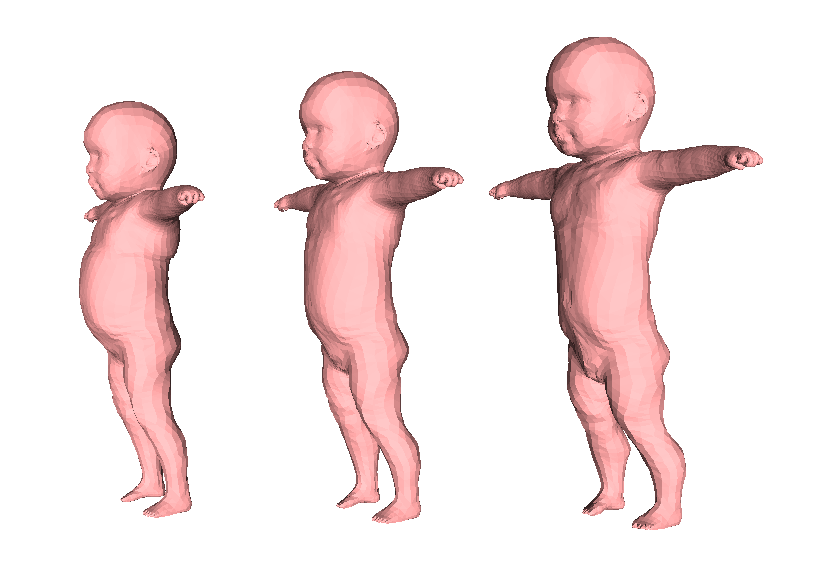}\llap{sc 2}
	\includegraphics[width=0.15\textwidth]{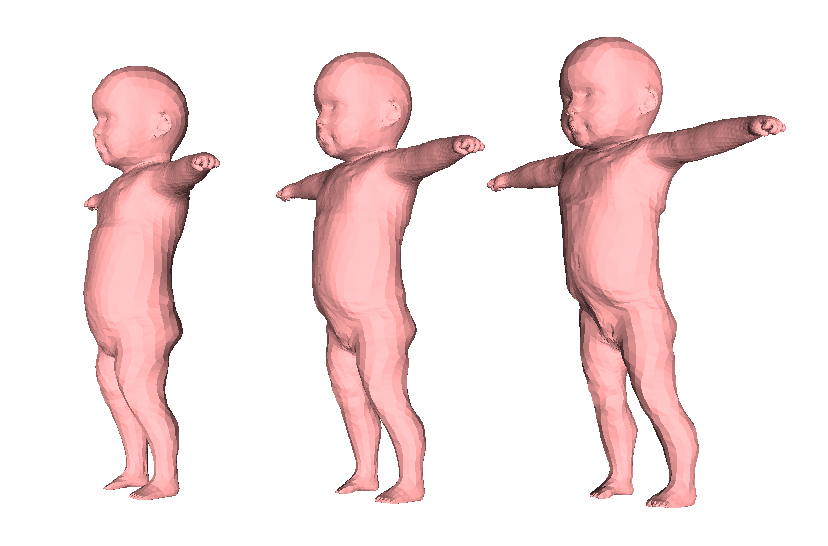}\llap{sc 3}
	\includegraphics[width=0.15\textwidth]{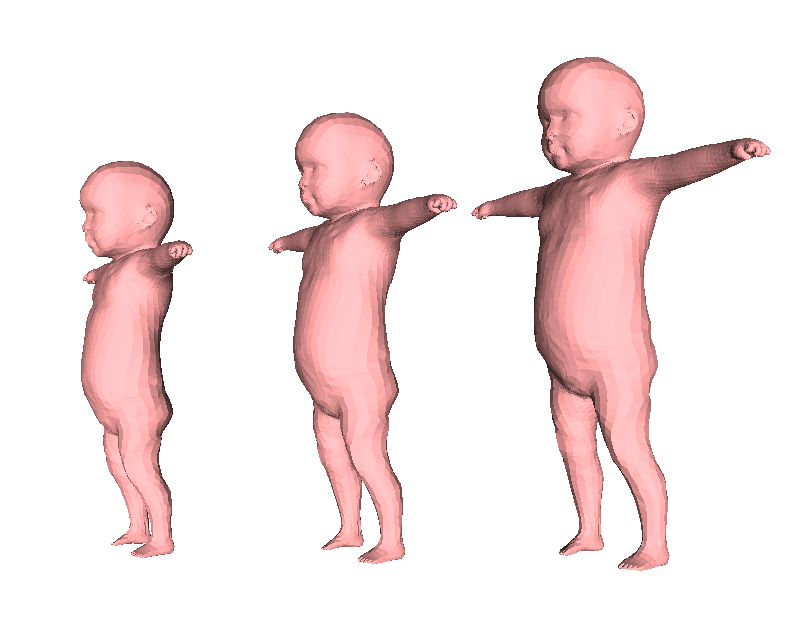}\llap{sc 1}
	\includegraphics[width=0.15\textwidth]{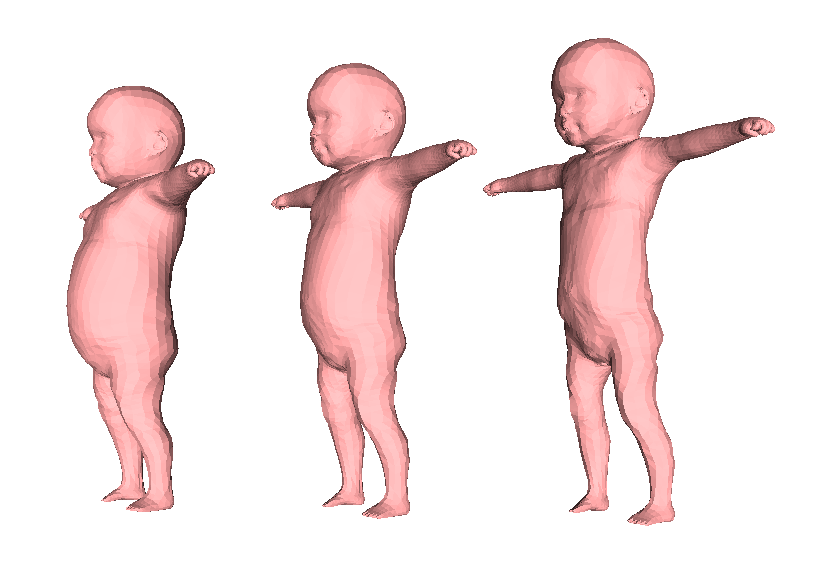}\llap{sc 2}
	\includegraphics[width=0.15\textwidth]{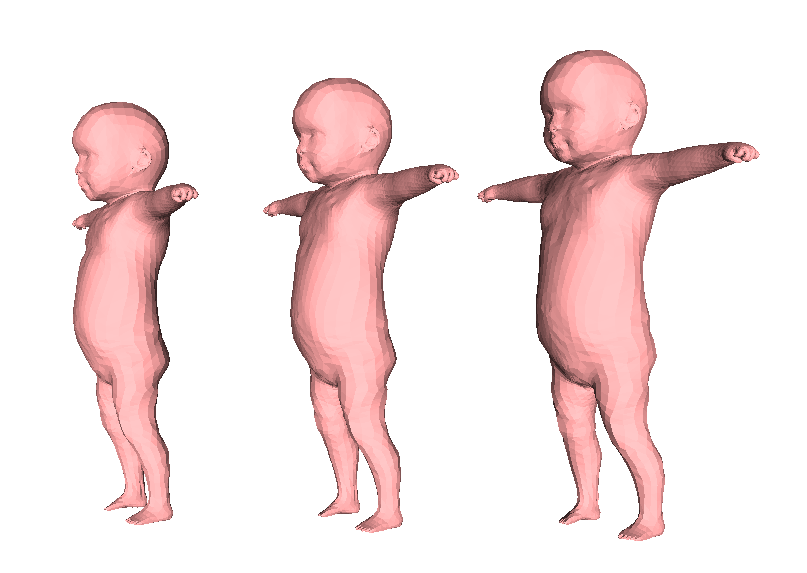}\llap{sc 3}
    \caption{First three shape principal components (sc). Top: SMIL,
      -2 to +2 standard deviations. Bottom: SMPL$_\mathrm{B}$, -0.5 to
      +0.5 standard deviations (i.e.~the adult shape space). The first components in the infant shape (SMIL sc 2 and 3) carry variation in trunk size / length, while the first components of SMPL$_\mathrm{B}$ show trunk variation mainly in the belly growing or shrinking.}
\label{fig:shapepc_smil}
\end{figure}

\begin{figure}[t]
	\centering
        \includegraphics[height=4.3cm]{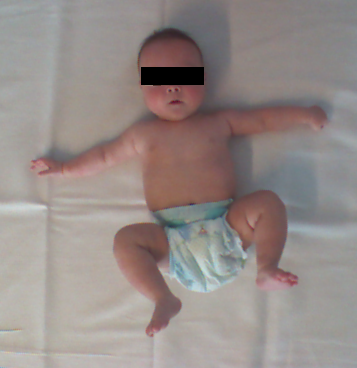}\llap{(a)}
        \includegraphics[height=4.3cm]{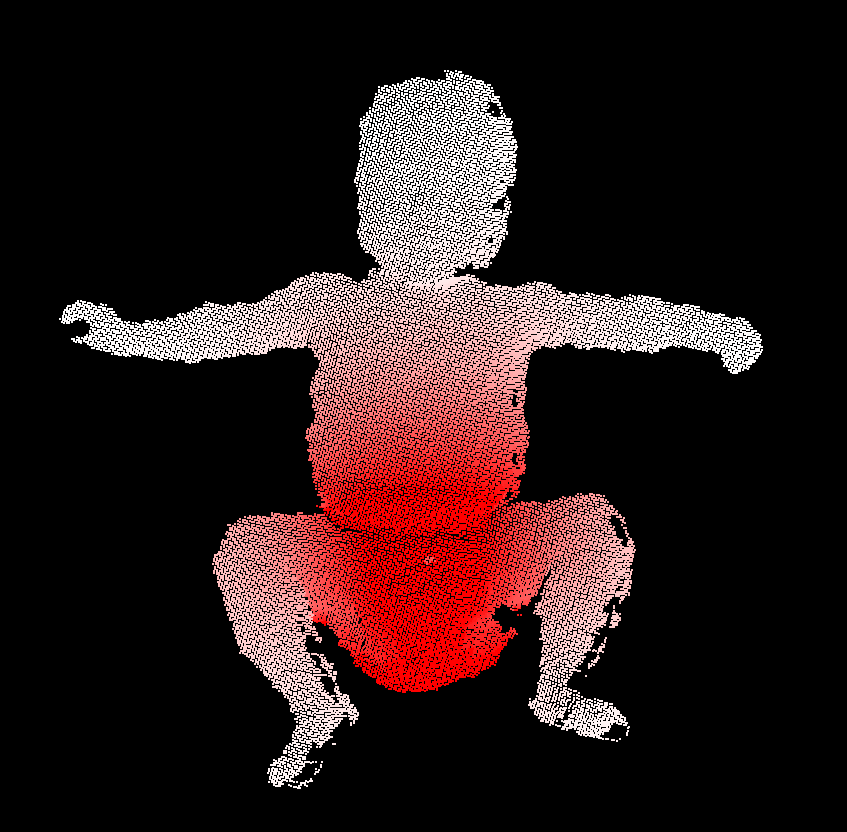}\llap{\textcolor{white}{(b)}}
    \caption{a) Original RGB image. b) Weights used for weighted PCA. White points have a high weight (value of 3), red point have a low weight (value of 1).
    The smooth transition is computed using the skin weights $W$.}
\label{fig:skinweights}
\end{figure}

We compute the new infant shape space by doing weighted principal component analysis (WPCA) on personalized shapes of all sequences.
Despite including the clothing segmentation in the creation of personalized shapes, clothing deformations can not be completely removed and  diapers typically tend to produce body shapes with an over-long trunk.
The recorded sequences contain infants with long-arm onesies, short-arm onesies, tights, diapers and without clothing.
These different clothing types cover different parts of the body.
As we want the shape space to be close to the real infant shape without clothing artifacts, we
use low weights for clothing points and high weights for skin points in the PCA.
The weights we use to train the model are:
3 for the scan points labeled as skin ($\mathrm{P_{skin}}$),
1 for the scan points labeled as clothing ($\mathrm{P_{cloth}}$),
and we compute smooth transition weights for the scan points near the cloth boundaries 
using the skin weights $W$ computed using the method in \cite{zhang2017detailed}.
Fig.~\ref{fig:skinweights} displays the weights used for the weighted PCA
on a sample frame.
We use the EMPCA algorithm\footnote{\url{https://github.com/jakevdp/wpca}}
computing weighted PCA with an iterative expectation-maximization approach.
We retain the first 20 shape components. We display the first 3 shape
components for SMIL and for the \smplb adult shape space in Fig.~\ref{fig:shapepc_smil}.

We create a pose data set by looping over all poses of all sequences and only add poses to the set if the dissimilarity to any pose in the set is larger than a threshold.
The new pose prior is learned from the final set containing 47K poses.
The final set contains 47K poses and is used to learn the new pose prior.
As the Gaussian pose prior can not penalize illegal poses, e.g. unnatural bending of knees,
we manually add penalties to avoid such poses.

The final SMIL model is composed of the shape space, the pose prior, and the base template, which is the mean of all personalized shapes.

\subsection{Manual Intervention}
In our method we use manual intervention three times: i) to decide which type of clothing the infant is wearing (see Sec~\ref{sec:data});
ii) to generate the initial model \smplb (see Sec~\ref{sec:inmo})
and iii) to define illegal poses in the pose prior.
The illegal poses are only defined once and the initial model is no longer used once SMIL is learned. However, given a new sequence, one still needs to manually define the type of clothing: short onesie, long onesie, tights, naked or diapers.
Each cloth type defines the corresponding number of color modes and priors to be used.
While this is the only remaining manual step in our method, we believe that a classifier predicting the clothing type from RGB images could be learned, making our method fully automatic.

\subsection{Method Parameters}
The values of the weights in the energy functions were empirically adjusted to keep the different terms balanced. 

For optimization of the main energy w.r.t.~shape and pose parameters (Eq.~\ref{eq:error_shape_pose}) and the modified energy w.r.t.~pose parameters (Eq.~\ref{eq:error_poseonly}) we use the weight values:
$\lambda_{\mathrm{skin}} = 800$,
$\lambda_{\mathrm{cloth}} = 300$,
$\lambda_{\mathrm{m2s}} = 400$,
$\lambda_{\mathrm{lm}} =1$,
$\lambda_{\mathrm{table}} = 10000$,
$\lambda_{\mathrm{sm}} = 800$,
$\lambda_{\mathrm{sc}} = 1$,
$\lambda_{\beta} = 1$ and
$\lambda_{\theta} = 0.15$.
%
For optimization of the energy w.r.t.~the model vertices (Eq.~\ref{eq:error_cpl}) we use the weight values:
$\lambda_{\mathrm{skin}} = 1000$,
$\lambda_{\mathrm{cloth}} = 500$,
$\lambda_{\mathrm{m2s}} = 1000$,
$\lambda_{\mathrm{lm}} = 0.03$,
$\lambda_{\mathrm{table}} = 10000$ and
$\lambda_{\mathrm{cpl}} = 1$.
For the creation of the personalized shape (Sec.~\ref{sec:pers}), we use weight values:
$\lambda_{\mathrm{skin}} = 100$,
$\lambda_{\mathrm{cloth}} = 100$
$\lambda_{\mathrm{\beta}} = 0.5$ and
$\lambda_{\mathrm{cpl}} = 0.4$.
Finally, for the initialization energy (Eq.~\ref{eq:init}), we use:
$\lambda_{\mathrm{j2d}} = 6$,
$\lambda_{\mathrm{\theta}} = 10$,
$\lambda_{\mathrm{a}} = 30$,
$\lambda_{\mathrm{\beta}} = 1000$,
$\lambda_{\mathrm{s2m}} = 30000$.
We keep the chosen weights constant for all experiments.

\section{Experiments}
\label{sec:exp}
As elaborated in the introduction, gathering high quality 3D scans of infants is highly unpractical, which is why we quantitatively evaluate SMIL and our initial model \smplb on the 37 acquired RGB-D sequences of infants.
We record the infants using a Microsoft Kinect V1, which is mounted 1 meter above an examination table, facing downwards.
All parents gave written informed consent for their child to participate in this study, which was approved by the ethics committee of Ludwig Maximilian University Munich (LMU).
The infants lie in supine position for three to five minutes without external stimulation, i.e.~there is no interaction with caregivers or toys.
The recorded infants are between 9 and 18 weeks of corrected age (post term), and 
their size range is 42 to 59 cm, with an average of 53.5 cm.
They wear different types of clothing: none, diaper, onesie shortarm / longarm, or tights.
All sequences together sum up to roughly 200K frames, and have an overall duration of over two hours.
We evaluate SMIL with a 9-fold cross-validation, using 33 sequences for training the shape space and the pose prior, and 4 for testing. We distribute different clothing styles across all training sets.
We measure the distance $\mathrm{E_{s2m}}$ (cf.~Eq.~\ref{eq:s2m}) of the scan to the model mesh by computing the Euclidean distance of each scan point to the mesh surface.
For evaluation, we consider all scan points to be labeled as skin, which reduces Eq.~\ref{eq:s2m} to Eq. \ref{eq:skin}. Note that we do not use the Geman-McClure function $\rho$ here, as we are interested in the actual Euclidean distances.

\begin{figure}[t]
	\centering
        \includegraphics[width=0.45\textwidth]{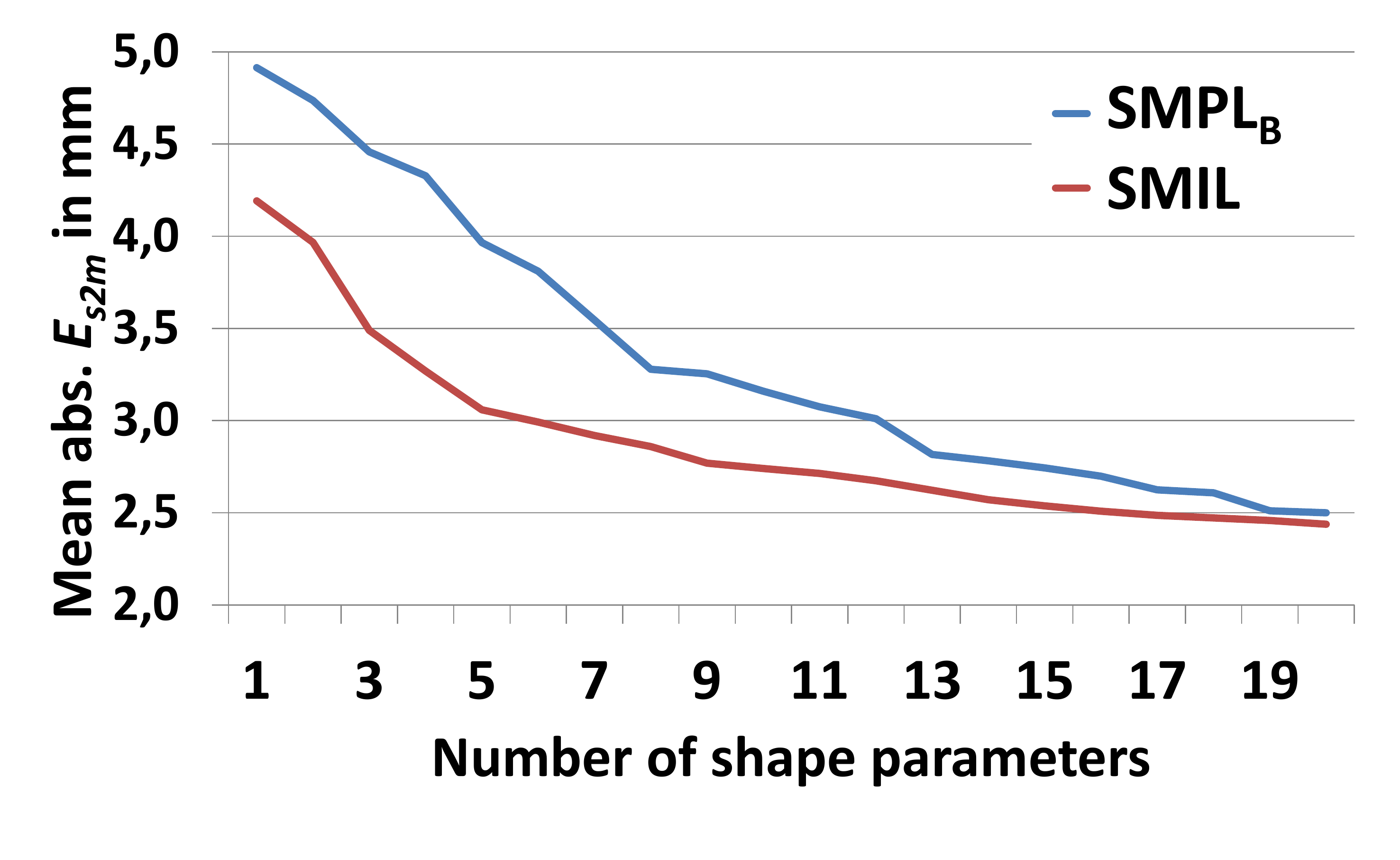}
    \caption{Average scan-to-mesh error Es2m in mm w.r.t. the number of shape parameters for the two models registered to all fusion scans.}
\label{fig:betaserror}
\end{figure}

\begin{figure}[t]
	\centering
	\includegraphics[width=0.11\textwidth]{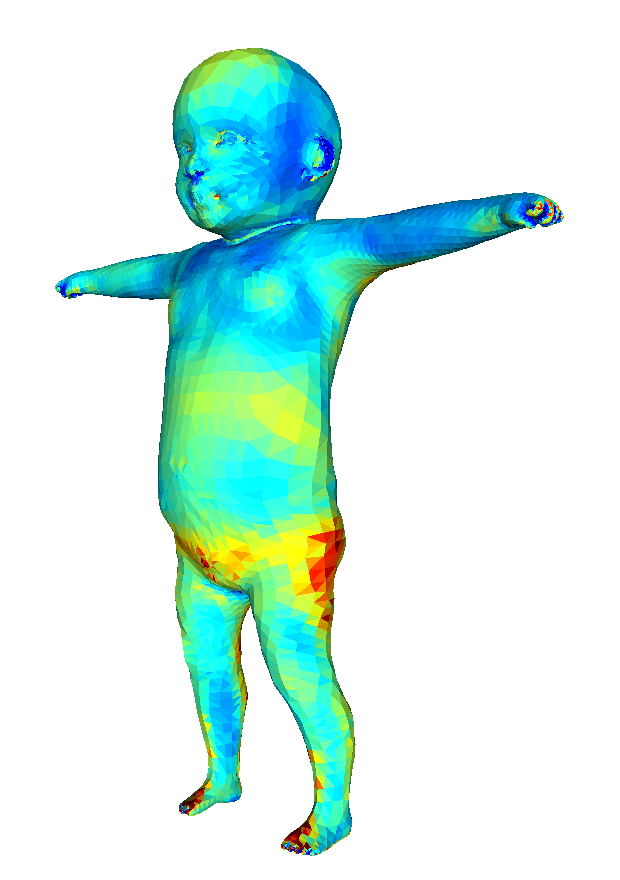}\llap{1 sc}
	\includegraphics[width=0.11\textwidth]{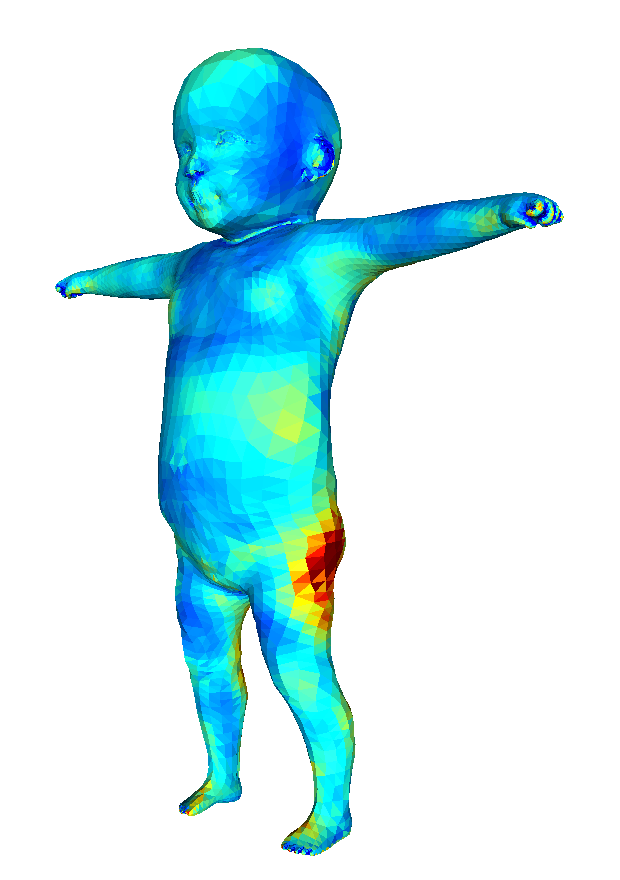}\llap{3 sc}
	\includegraphics[width=0.11\textwidth]{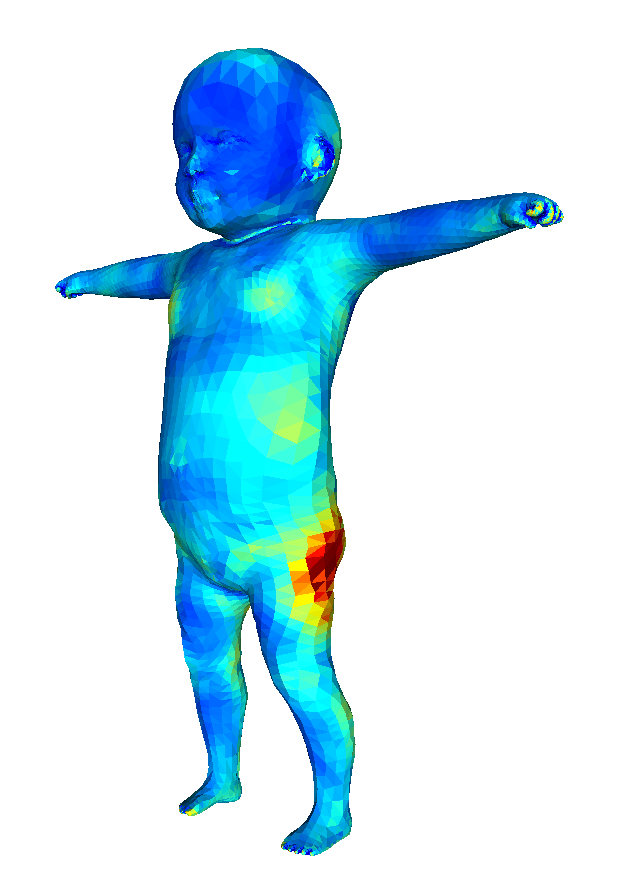}\llap{5 sc}
	\includegraphics[width=0.11\textwidth]{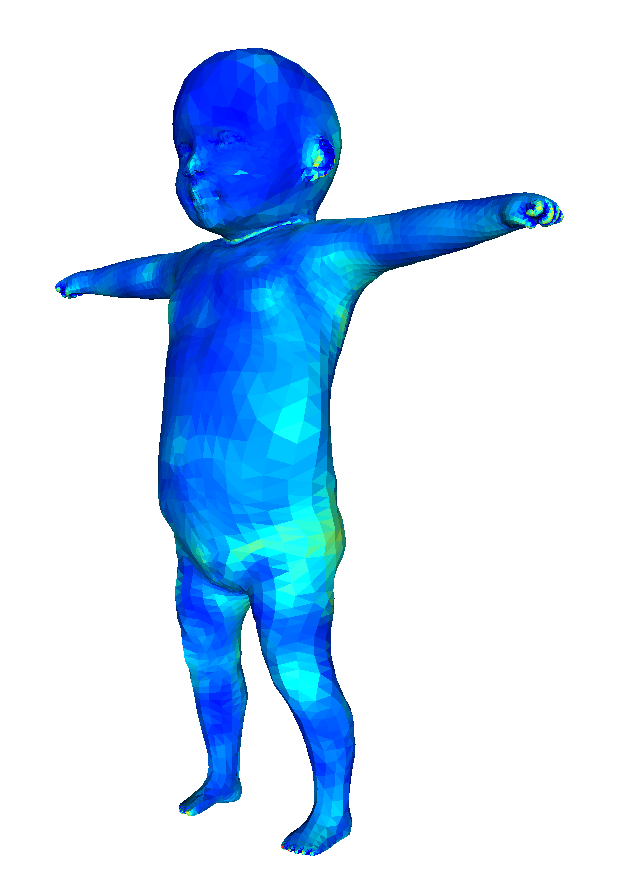}\llap{20 sc}
	\includegraphics[width=0.11\textwidth]{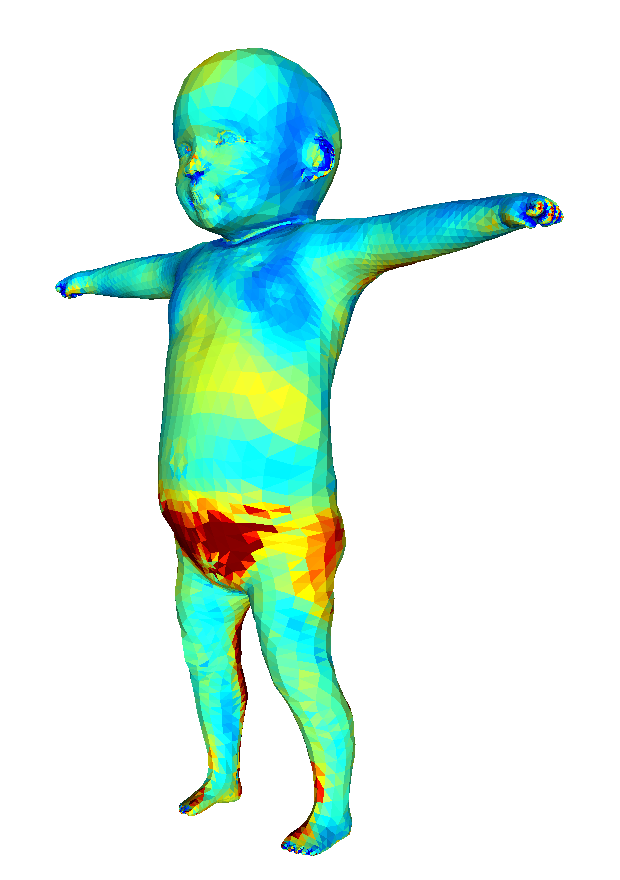}\llap{1 sc}
	\includegraphics[width=0.11\textwidth]{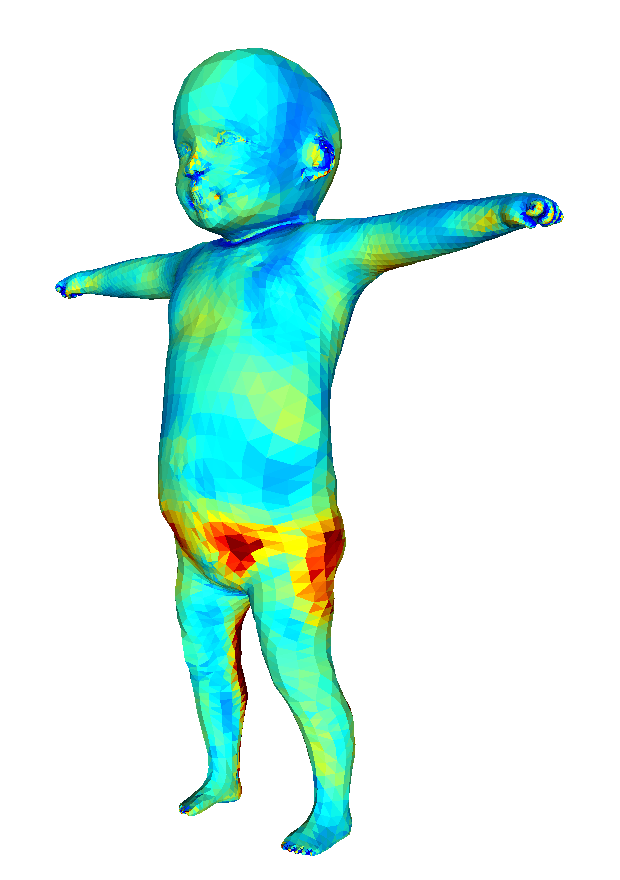}\llap{3 sc}
	\includegraphics[width=0.11\textwidth]{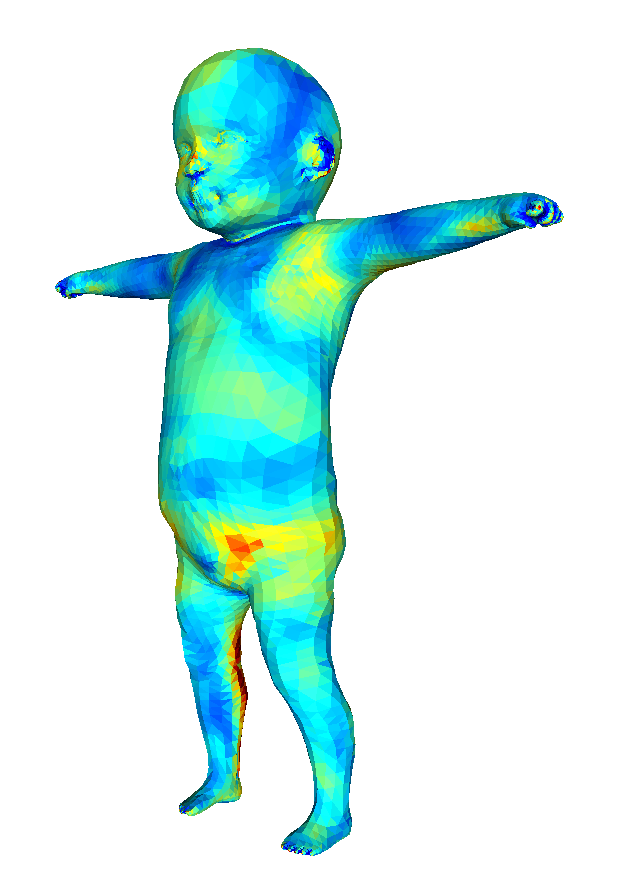}\llap{5 sc}
	\includegraphics[width=0.11\textwidth]{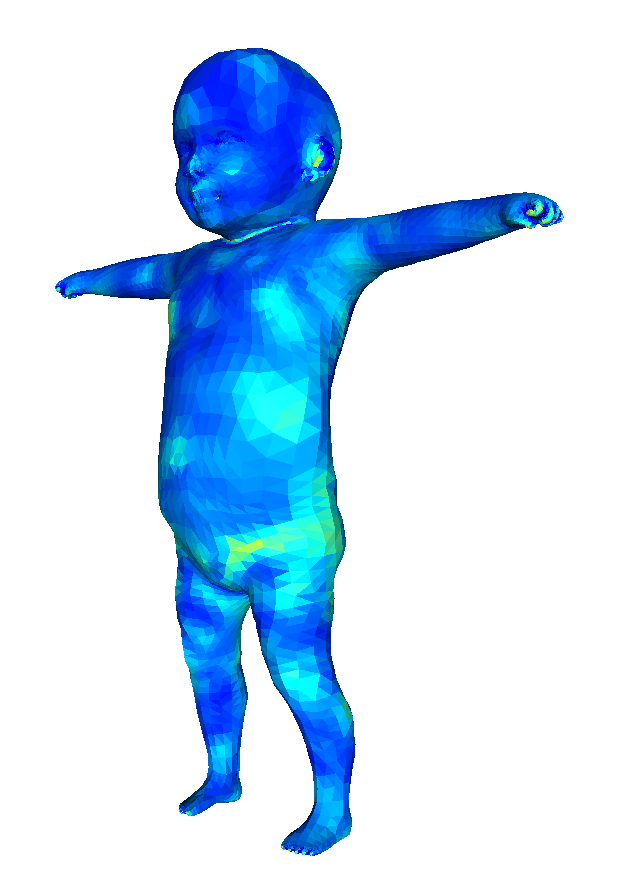}\llap{20 sc}
    \caption{Average error heatmaps for SMIL and \smplb on fusion clouds for different numbers of shape components (sc). Top: SMIL. Bottom: SMPL$_\mathrm{B}$. Blue means 0 mm, red means $\geq$ 10 mm.}
\label{fig:fusion_heat}
\end{figure}

To compare the \smplb shape space to the SMIL shape space, we register both models to each of the 37 fusion scans,
using different numbers of shape components.
Results are displayed in Fig.~\ref{fig:betaserror}.
We plot average error heatmaps for using the first 1, 3, 5 and all 20 shape components for the registrations (Fig.~\ref{fig:fusion_heat}).
We observe lower error for SMIL for smaller numbers of shape parameters, and a nearly identical error when using all 20 parameters.
Note: \smplb is not the SMPL model \cite{loper2015smpl}, but our initial infant model, registered to the data using our method.

To evaluate how well the computed personalized shapes and poses explain the input sequences, we calculate $\mathrm{E_{s2m}}$ for all 200K frames.
SMIL achieves an average scan-to-mesh distance of 2.51 mm (SD 0.21 mm), \smplb has an average $\mathrm{E_{s2m}}$ of 2.67 mm (SD 0.22 mm)

\begin{figure}[t]
\centering
	\includegraphics[width=0.49\textwidth]{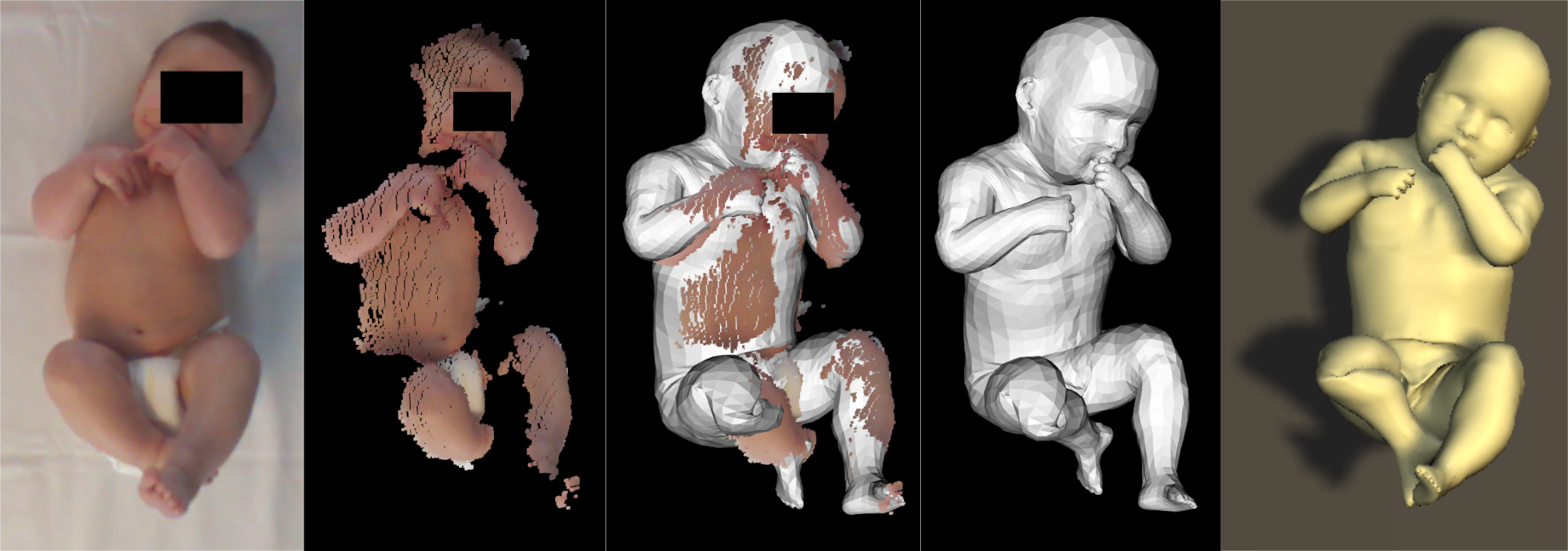}
	\includegraphics[width=0.49\textwidth]{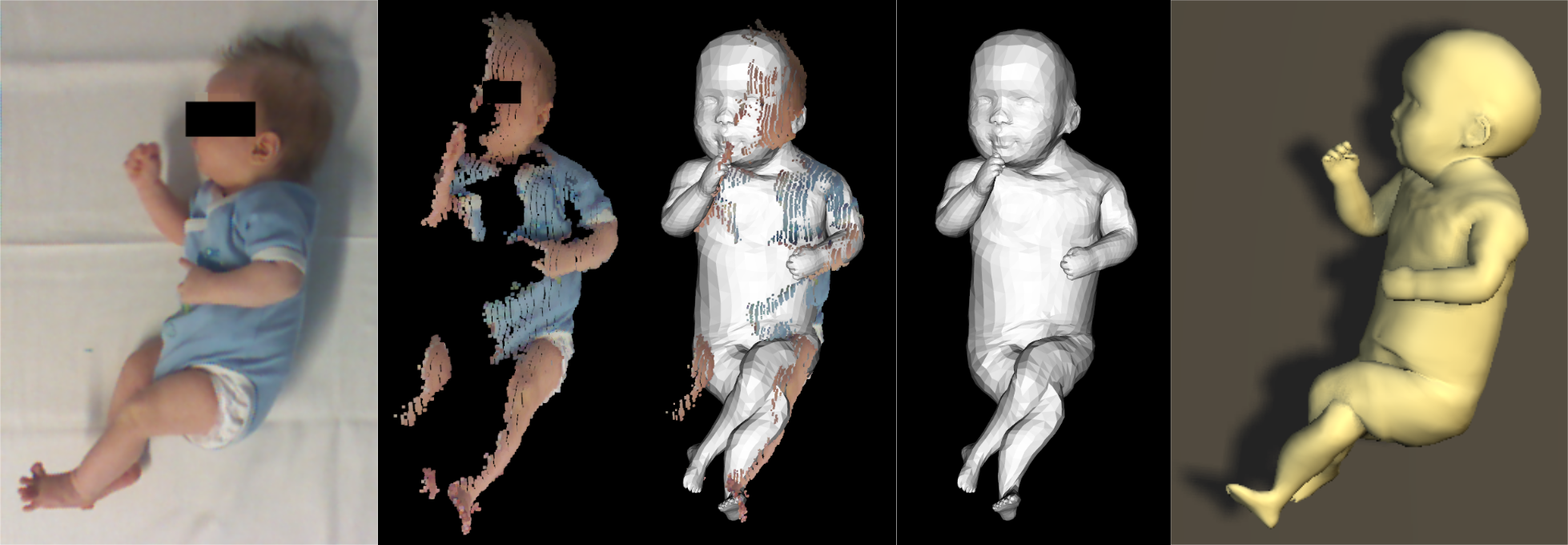}
	\caption{Top: unnatural pose sample. Bottom: Failure case sample. From left to right: RGB image, 3D point cloud (rotated for improved visibility), overlay of point cloud and registration result, registration result, rendered result from same viewpoint as RGB image.}
\label{fig:fail1}
\end{figure}

Due to the lack of ground truth data for evaluation of infant pose correctness,
we perform a manual inspection of all sequences to reveal pose errors.
We distinguish between ``unnatural poses'' and ``failure cases''. Unnatural poses contain errors in pose, like implausible rotations of a leg (cf.~Fig.~\ref{fig:fail1} top row), while the overall registration is plausible, i.e. the 3D joint positions are still at roughly the correct position.
Failure cases denote situations in which the optimization gets stuck in a local minimum with a clearly wrong pose, i.e.~one model body part registered to a scan part which it does not belong to (cf.~Fig.~\ref{fig:fail1} bottom row).
We count 16 unnatural leg/foot rotations lasting 41 seconds (= 0.54\% of roughly 2 hours) and 18 failure cases (in 7 sequences) lasting 49 seconds (= 0.66\% of roughly 2 hours).

\begin{figure}[t]
	\centering
	\includegraphics[height=2.85cm]{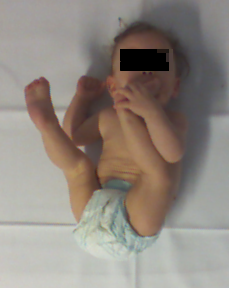}\llap{(a)}
	\includegraphics[height=2.85cm]{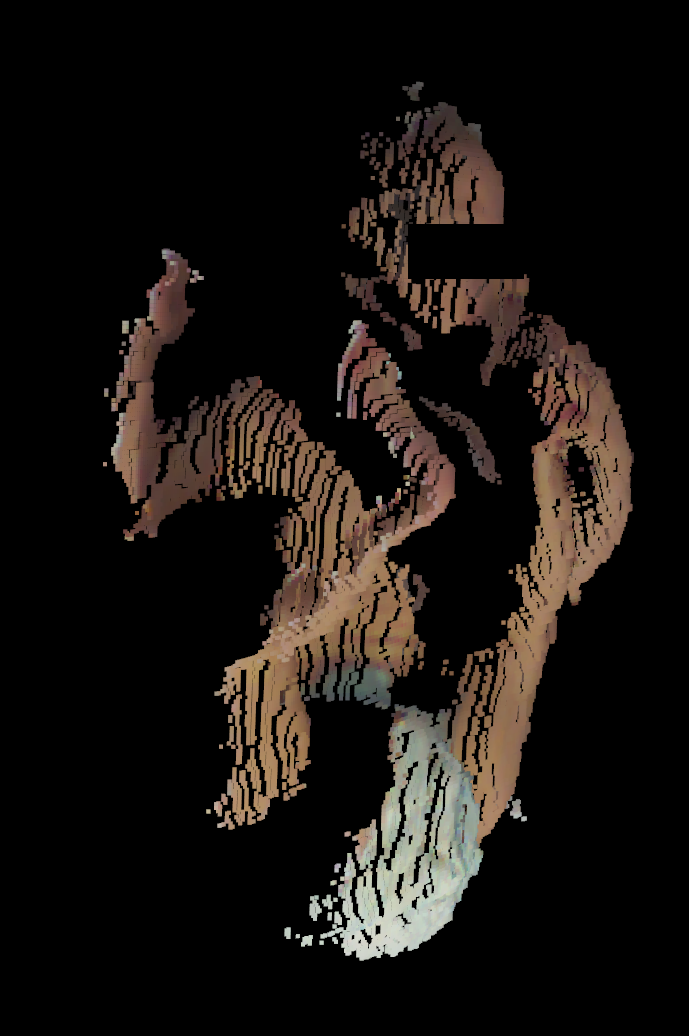}\llap{\color{white}{(b)}}
	\includegraphics[height=2.85cm]{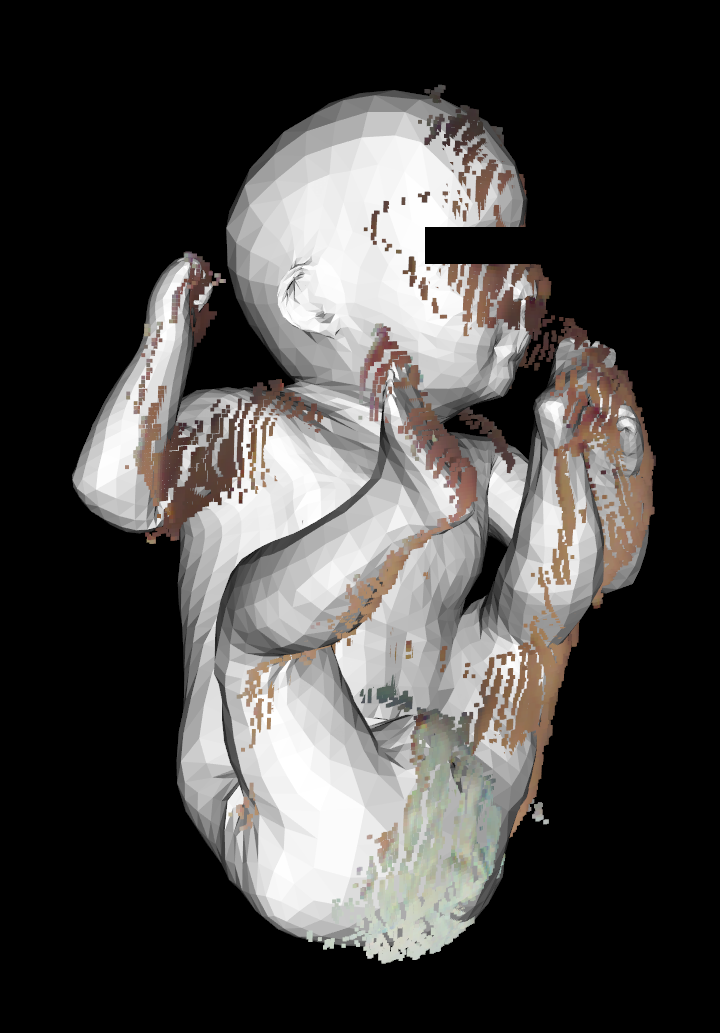}\llap{\color{white}{(c)}}
	\includegraphics[height=2.85cm]{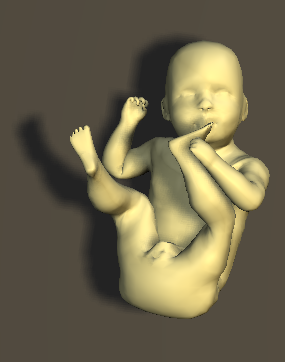}\llap{\color{white}{(d)}}
    \caption{Older infant in very challenging pose. (a) RGB image, (b) 3D point cloud (rotated for improved visibility), (c) overlay of point cloud and SMIL registration result, (d) rendered SMIL registration result.}
\label{fig:older}
\end{figure}

To evaluate how well SMIL generalizes to older infants, we register the model to 25 sequences of infants at the age between 21 and 36 weeks, at an average of 26 weeks.
The resulting average scan to mesh distance is 2.83 mm (SD: 0.31 mm).
With increasing age, infants learn to perform directed movements, like touching their hands, face, or feet, as displayed in Fig.~\ref{fig:older}.
This makes motion capture even more challenging, as standard marker-based methods would not be recommended because of the risk of infants grabbing (and possibly swallowing) markers.

\textbf{Failure cases.}
The most common failure is a mixup of feet, i.e.~left  foot of the model registered to the right foot of the scan and vice versa.
Despite our energy having the interpenetration penalty $\mathrm{E}_{sc}$, 
we observe a few cases where the legs interpenetrate, as in the bottom row in Fig.~\ref{fig:fail1}.
The registration of all sequences is time consuming (between 10 and 30 seconds per frame),
so rerunning the full 200K registrations many times to optimize the parameters is
not feasible.
The energy term weights are manually selected in order to balance the different terms, and by visually inspecting the results of some sequences.
Further manual adjustment of the $\mathrm{E}_{sc}$ weight could fix these rare cases.
In the example in the top row of Fig.\ref{fig:fail1}, the right knee is twisted in an unnatural way after the right foot was completely occluded.
When the foot is visible again, the pose recovers (knee twisted for 5-6 seconds).
Similar to the first failure case, a higher weight on the pose prior
would prevent such cases, but finding the perfect weight which
completely forbids all illegal poses while allowing all legal poses
would require a significant engineering effort or more training data.

\textbf{Motion analysis case study.}
To show that SMIL captures enough motion information for medical assessment we conduct a case study on GMA.
Two trained and certified GMA-experts perform GMA in different videos. 
We use five stimuli: 
i) the original RGB videos (denoted by $\mathrm{V_{rgb}}$), and
ii) the synthetic registration videos ($\mathrm{V_{reg}}$).
For the next three stimuli we use the acquired poses of infants,
but we animate a body with a different shape, namely
iii) a randomly selected shape of another infant ($\mathrm{V_{other}}$), 
iv) an extreme shape producing a very thick and large baby ($\mathrm{V_{large}}$),
and v) the mean shape ($\mathrm{V_{mean}}$).
We exclude three of the 37 sequences, as two are too short and one has non-nutritive sucking, making it non suitable for GMA.
As the number of videos to rate is high (34*5), for iv) and v) we only use 50\% of the sequences, resulting in 136 videos.
For a finer evaluation, we augment GMA classes \textit{definitely abnormal} (DA), \textit{mildly abnormal} (MA), \textit{normal suboptimal} (NS), and \textit{normal optimal} (NO) of \cite{hadders2004general} into a one to ten scale. Scores 1-3 correspond to DA, 4-5 to MA, 6-7 to NS, and 8-10 to NO.
We consider two ratings with an absolute difference $\leq 1$ to {\it agree}, and otherwise to {\it disagree}.

\begin{figure}[t]
\centering
	\includegraphics[width=0.49\textwidth]{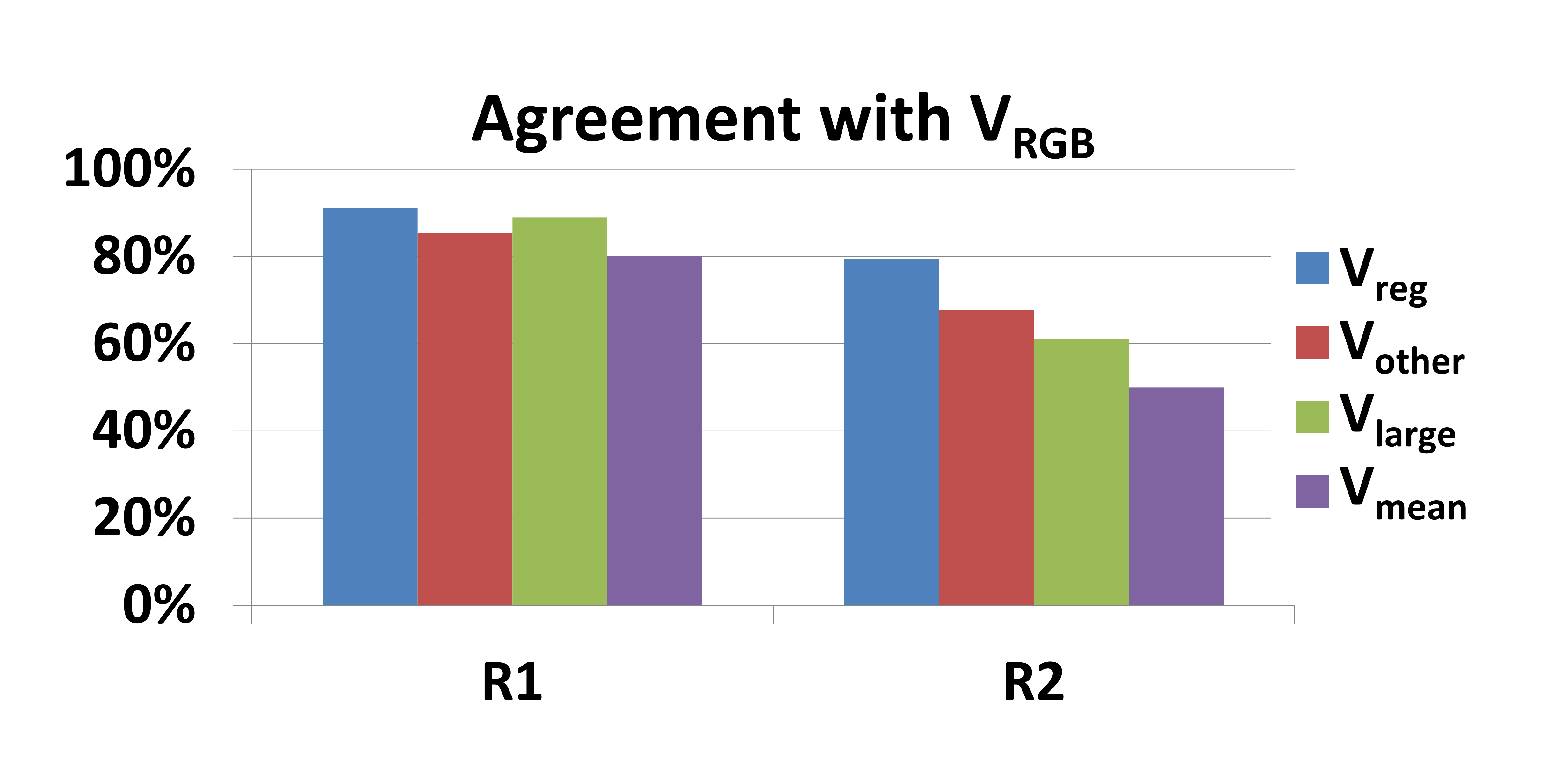}
	\caption{Results of GMA case study. Percentage of ratings of synthetic sequences, generated using SMIL, that {\it agree} with the reference ratings 
	$\mathrm{R_1V_{rgb}}$ (left) and $\mathrm{R_2V_{rgb}}$ (right), respectively.
	 $\mathrm{V_{\{reg, other, large, mean\}}}$ denotes different stimuli.}
\label{fig:bins}
\end{figure}

Rater $\mathrm{R_1}$ is a long-time GMA teacher and has worked on GMA for over 25 years, $\mathrm{R_2}$ has 15 years experience in GMA.
Average rating score (and standard deviation) for R1 is 4.7 (1.4), for R2 4.0 (1.9).
The agreement on original RGB ratings $\mathrm{V_{rgb}}$ between
$\mathrm{R_1}$ and $\mathrm{R_2}$ is 65\%.
This further stresses that GMA is challenging and its automation important.
In Fig.~\ref{fig:bins}, we present rating differences between synthetic and reference sequences. Each rater is compared to her own $\mathrm{V_{rgb}}$ ratings as a reference.
$\mathrm{R_1V_{reg}}$ ratings {\it agree} on 91\% of the reference ratings, whereas $\mathrm{R_2}$ achieves an agreement rate of 79\%.
The agreement decreases more ($\mathrm{R_2}$) or less ($\mathrm{R_1}$) when 
the motions are presented with a different body shape.
We intend to conduct further studies to elucidate the biases introduced by the variation of the infants' shape.

\begin{figure}[t]
\centering
\includegraphics[height=4cm]{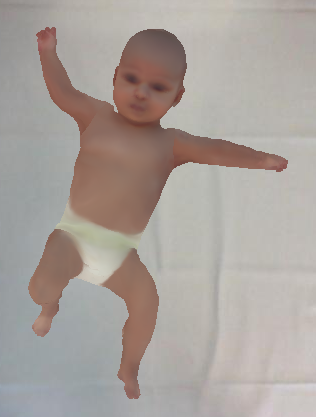}\llap{(a) }
\includegraphics[height=4cm]{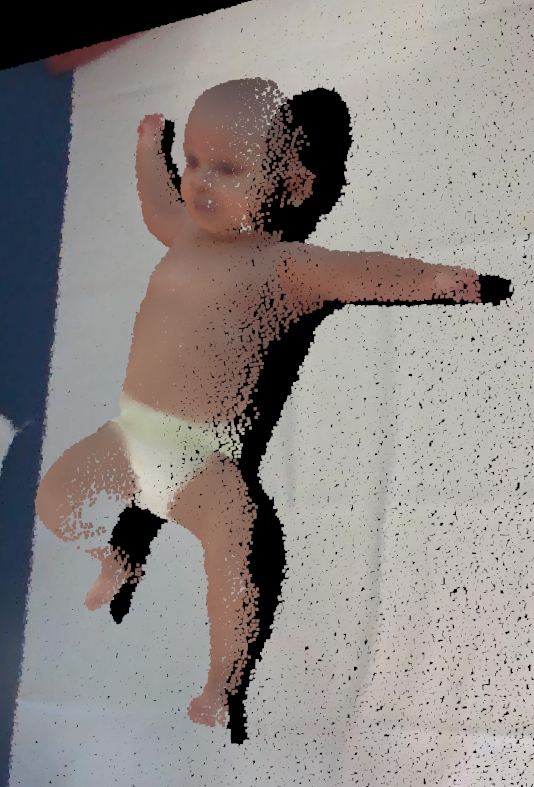}\llap{(b) }
\includegraphics[height=4cm]{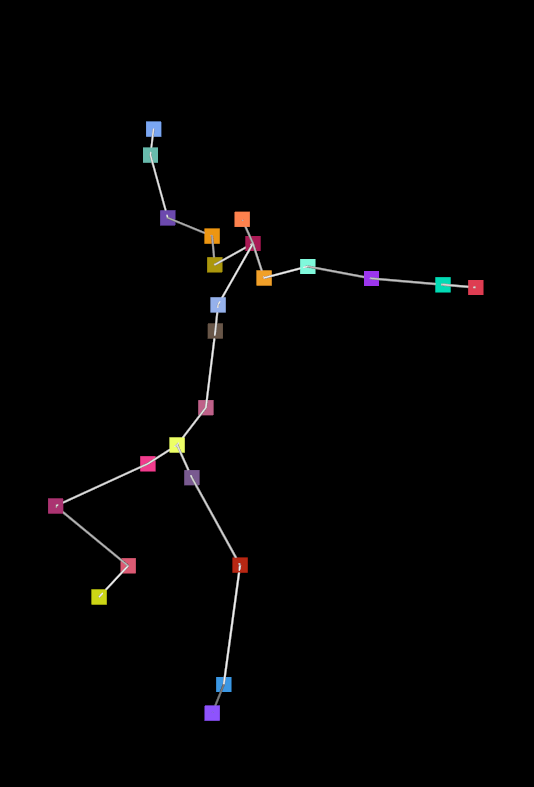}\llap{\color{white}{(c) }}
\includegraphics[height=2.9cm]{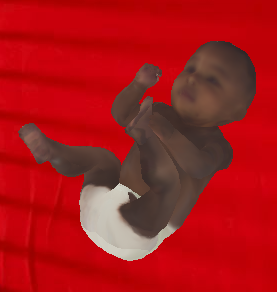}\llap{\color{white}{(d) }}
\includegraphics[height=2.9cm]{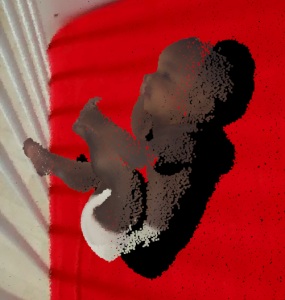}\llap{\color{white}{(e) }}
\includegraphics[height=2.9cm]{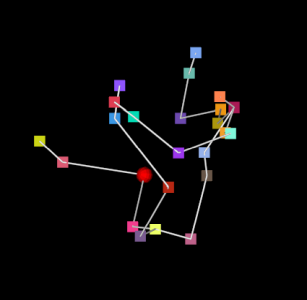}\llap{\color{white}{(f) }}
\caption{Two data samples created using SMIL containing: RGB image (a,d); point cloud from depth image (b, e) and ground truth skeleton (c, f). Viewpoint for (b), (c), (e), and (f) is slightly rotated to the side.}
\label{fig:minirgbd}
\end{figure}

\textbf{Generation of realistic data.}
Human body models have been used to create training data for deep neural networks \cite{varol2017learning}.
We used SMIL to create a realistic (but yet privacy preserving) data set of \textit{Moving INfants In RGB-D} (MINI-RGBD) \cite{hesse2018computer}, which is available at \url{http://s.fhg.de/mini-rgbd}.
To create the data set, we captured shape and pose of infants from RGB-D sequences as described in Sec.~\ref{sec:reg}, but additionally captured texture.
We selected random subsets of shapes and textures, and averaged them to create new, synthetic, but realistic shapes and textures.
We mapped the real captured poses to the new synthetic infants and extracted ground truth 3D joint positions.
We used OpenDR \cite{loper2014opendr} for rendering RGB and depth images to resemble commodity RGB-D sensors.
We created the data set with the intention to provide an evaluation set for pose estimation in medical infant motion analysis scenarios.
A sample of the data is displayed in Fig.~\ref{fig:minirgbd}.
We have showed that SMIL can be used to create realistic RGB-D data, and we plan to create a larger data set to provide enough data to train neural networks for infant shape and pose estimation.



\section{Discussion / Conclusion}
\textit{Why does it work?}
Even though each RGB-D frame is a partial observation of the shape, 
the motions in the different frames of a sequence reveal previously hidden body parts.
Moreover, even though the backs of the infants are rarely visible, we can still faithfully infer where the infants' backs are by taking into account the background table constrains.
This lets us accumulate shape information over sequences.
In addition, we leverage 2D pose estimation from RGB in two ways: i) it allows us to add landmark constraints to guide the model where depth-only approaches fail and 
ii) allows us to find a good initialization frame and circumvent the need of predefined poses.

\textit{Why has it not been done before?}
Most previous work has been done on adults, who can be instructed and 3D scanned much more easily.
Recording infants poses more challenges (see Sec.~\ref{sec:data}).
As our setup only relies on a low-cost RGB-D camera and a laptop, we can capture infants with great flexibility,
for instance at a children's hospital that parents and children are visiting anyway.
The creation of an initial model was not straightforward.
Thanks to the flexibility of the SMPL model, we were able to adapt it to infants and get a starting point for capturing real infant shapes.

\textbf{Limitations.}
Due to the data resolution, fine structures such as fingers and toe movements are not captured.
Moreover, the model is not yet able to capture facial expressions.
In our registration results, we found several cases of unusual neck twists.
The SMIL neck seems to be longer than the average infant neck, which is why it is sometimes twists to achieve a compression and match the data.

\textbf{Conclusion.}
We contribute a method for learning a body model from RGB-D data of freely moving infants.
We show that our learned Skinned Multi-Infant Linear model (SMIL) factorizes pose and shape and achieves metric accuracy of 2.5 mm.
We further applied the model to the task of medical infant motion analysis.
Two expert GMA raters achieve a scoring agreement of 91\%, respectively 79\%, when comparing the assessment of movement quality from standard RGB video and from rendered SMIL registration results of the same sequence.
Our method is a step towards a fully automated system for GMA, to help early-detect neurodevelopmental disorders like cerebral palsy.

\textbf{Future work.}
In this work we have not learned pose-dependent shape deformations for infants, and we reused the scaled down pose blend shapes of SMPL. 
While numerically these provide sufficient accuracy, 
 we will learn infant pose blend shapes to further increase the realism of SMIL.

In an ongoing study, we are collecting more RGB-D data by taking advantage of the lightweight recording setup: we capture the infants' motions in their homes to minimize stress and effort for both infants and parents.
We will further pursue the clinical goal to create SMIL: the automation of GMA, i.e. learn how to infer GMA ratings from captured motions.
We are currently applying the system to infants affected by spinal muscular atrophy, 
with the goal to quantify the disease progress as well as the impact of therapy.


%

%

\ifCLASSOPTIONcompsoc
  \section*{Acknowledgments}
\else
  \section*{Acknowledgment}
\fi

Authors thank
Javier Romero for helpful discussions,
Mijna Hadders-Algra and Uta Tacke for GMA ratings and medical insights, 
R. Weinberger for data acquisition and help with evaluation
and Paul Soubiran for the video voice over.

Disclosure: MJB has received research gift funds from Intel, Nvidia, Adobe, Facebook,
and Amazon. While MJB is a part-time employee of Amazon, his 
contribution was performed solely at, and funded solely by, MPI.

\ifCLASSOPTIONcaptionsoff
  \newpage
\fi



\bibliographystyle{IEEEtran}
\end{document}